\documentclass[10pt,twocolumn,letterpaper]{article}

\usepackage{iccv}
\usepackage{times}
\usepackage{epsfig}
\usepackage{graphicx}
\usepackage{amsmath}
\usepackage{amssymb}
\usepackage{lipsum}
\usepackage{marvosym}

\usepackage[pagebackref=true,breaklinks=true,letterpaper=true,colorlinks,bookmarks=false]{hyperref}
\hypersetup{citecolor=[RGB]{119,185,0}}

\iccvfinalcopy 

\def\iccvPaperID{7009} 

\ificcvfinal\fi

\usepackage[mathscr]{eucal}
\usepackage{amssymb,amsmath,amsfonts,latexsym}
\usepackage{amsmath,graphicx,bm,xcolor,url}
\usepackage[caption=false]{subfig} 
\usepackage{bm}
\usepackage{mathrsfs}
 \usepackage{amsthm}

\catcode`~=11 \def\UrlSpecials{\do\~{\kern -.15em\lower .7ex\hbox{~}\kern .04em}} \catcode`~=13 

\allowdisplaybreaks[3]

\DeclareMathAlphabet{\mathbsf}{OT1}{cmss}{bx}{n}
\DeclareMathAlphabet{\mathssf}{OT1}{cmss}{m}{sl}

\DeclareSymbolFont{bsfletters}{OT1}{cmss}{bx}{n}  
\DeclareSymbolFont{ssfletters}{OT1}{cmss}{m}{n}
\DeclareMathSymbol{\bsfGamma}{0}{bsfletters}{'000}
\DeclareMathSymbol{\ssfGamma}{0}{ssfletters}{'000}
\DeclareMathSymbol{\bsfDelta}{0}{bsfletters}{'001}
\DeclareMathSymbol{\ssfDelta}{0}{ssfletters}{'001}
\DeclareMathSymbol{\bsfTheta}{0}{bsfletters}{'002}
\DeclareMathSymbol{\ssfTheta}{0}{ssfletters}{'002}
\DeclareMathSymbol{\bsfLambda}{0}{bsfletters}{'003}
\DeclareMathSymbol{\ssfLambda}{0}{ssfletters}{'003}
\DeclareMathSymbol{\bsfXi}{0}{bsfletters}{'004}
\DeclareMathSymbol{\ssfXi}{0}{ssfletters}{'004}
\DeclareMathSymbol{\bsfPi}{0}{bsfletters}{'005}
\DeclareMathSymbol{\ssfPi}{0}{ssfletters}{'005}
\DeclareMathSymbol{\bsfSigma}{0}{bsfletters}{'006}
\DeclareMathSymbol{\ssfSigma}{0}{ssfletters}{'006}
\DeclareMathSymbol{\bsfUpsilon}{0}{bsfletters}{'007}
\DeclareMathSymbol{\ssfUpsilon}{0}{ssfletters}{'007}
\DeclareMathSymbol{\bsfPhi}{0}{bsfletters}{'010}
\DeclareMathSymbol{\ssfPhi}{0}{ssfletters}{'010}
\DeclareMathSymbol{\bsfPsi}{0}{bsfletters}{'011}
\DeclareMathSymbol{\ssfPsi}{0}{ssfletters}{'011}
\DeclareMathSymbol{\bsfOmega}{0}{bsfletters}{'012}
\DeclareMathSymbol{\ssfOmega}{0}{ssfletters}{'012}

\newcommand{\qednew}{\nobreak \ifvmode \relax \else
      \ifdim\lastskip<1.5em \hskip-\lastskip
      \hskip1.5em plus0em minus0.5em \fi \nobreak
      \vrule height0.75em width0.5em depth0.25em\fi}

\usepackage{hyperref}
\usepackage{stfloats}
\usepackage{subfig}
\usepackage{xcolor}
\usepackage{makecell}
\usepackage{xspace}
\usepackage{url}
\usepackage{wrapfig}
\usepackage{multirow}
\usepackage{makecell}
\usepackage{booktabs, colortbl}
\usepackage{longtable}
\usepackage{hhline}
\usepackage{color}
\usepackage{algpseudocode}
\usepackage{listings}
\usepackage[ruled]{algorithm2e}

\definecolor{myblue}{HTML}{0072C6}
\definecolor{myyellow}{HTML}{FFFADF}
\definecolor{myred}{HTML}{FF0000}

\usepackage[capitalize,noabbrev]{cleveref}
\usepackage{appendix}
\usepackage{cleveref}
\crefname{section}{Section}{Sections}
\crefname{theorem}{Theorem}{Theorems}
\crefname{lemma}{Lemma}{Lemmas}
\crefname{equation}{Equation}{Equations}
\crefname{proposition}{Proposition}{Propositions}
\crefname{claim}{Claim}{Claims}
\crefname{appendix}{Appendix}{Appendices}
\crefname{algorithm}{Algorithm}{Algorithms}
\crefname{figure}{Figure}{Figs}
\crefname{table}{Table}{Tables}
\crefname{remark}{Remark}{Remarks}
\crefname{definition}{Definition}{Definitions}
\crefname{equation}{Equation}{Equations}
\crefname{corollary}{Corollary}{Corollaries}
\crefalias{section}{appendix}

\definecolor{codegreen}{rgb}{0,0.6,0}
\definecolor{codegray}{rgb}{0.5,0.5,0.5}
\definecolor{codepurple}{rgb}{0.58,0,0.82}
\definecolor{backcolour}{rgb}{0.95,0.95,0.92}
\definecolor{lightgreen}{HTML}{30E1C8}
\definecolor{lightblue}{HTML}{0254D6}
\definecolor{cite_color}{HTML}{114083}
\definecolor{link_color}{RGB}{153, 0,0}  
\definecolor{url_color}{RGB}{153, 102,  0}
\definecolor{emp_color}{RGB}{0,0,255}

\usepackage{xspace}
\newcommand{\ours}[0]{DDP\xspace}

\newcommand{\tablestyle}[2]{\setlength{\tabcolsep}
{#1}\renewcommand{\arraystretch}{#2}\centering\small}
\newlength\savewidth
\begin{document}
\definecolor{baselinecolor}{gray}{.9}
\newcommand{\baseline}[1]{\cellcolor{baselinecolor}{#1}}
\newcommand{\gray}[1]{\textcolor{gray}{#1}}
\definecolor{mygray}{gray}{.8}

\makeatletter
\renewcommand{\paragraph}{%
  \@startsection{paragraph}{4}%
  {\z@}{0.2\baselineskip}{-1em}{\normalfont\normalsize\bfseries}%
}
\makeatother

\newcommand{\algcomment}[1]{%
    \vspace{-\baselineskip}%
    \noindent%
    {\footnotesize #1\par}%
    \vspace{\baselineskip}%
}
\renewcommand{\thefootnote}{}

\title{DDP: Diffusion Model for Dense Visual Prediction}

\author{
    Yuanfeng Ji$^{1*}$, 
    Zhe Chen$^{3*}$,
    Enze Xie$^{2\dagger}$,
    Lanqing Hong$^{2}$, 
    Xihui Liu$^{1}$, \\
    Zhaoqiang Liu$^{2}$,
    Tong Lu$^{3}$,
    Zhenguo Li$^{2}$,
    Ping Luo$^{1}$\\
    $^1$The University of Hong Kong~~~
    $^2$Huawei Noah’s Ark Lab~~~
    $^3$Nanjing University~~~ \\
    {\small \url{https://github.com/JiYuanFeng/DDP}}
}

\maketitle

\ificcvfinal\thispagestyle{empty}\fi

\footnotetext{$*$ Equal contribution.}
\footnotetext{$\dagger$ Corresponding author.}

\begin{abstract}
We propose a simple, efficient, yet powerful framework for dense visual predictions based on the conditional diffusion pipeline.
Our approach follows a ``noise-to-map" generative paradigm for prediction by progressively removing noise from a random Gaussian distribution, guided by the image.
The method, called DDP, efficiently extends the denoising diffusion process into the modern perception pipeline.
Without task-specific design and architecture customization, \ours is easy to generalize to most dense prediction tasks, e.g., semantic segmentation and depth estimation.
In addition, \ours shows attractive properties such as dynamic inference and uncertainty awareness, in contrast to previous single-step discriminative methods.
We show top results on three representative tasks with six diverse benchmarks, without tricks, \ours achieves state-of-the-art or competitive performance on each task compared to the specialist counterparts.
For example, semantic segmentation (83.9 mIoU on Cityscapes), BEV map segmentation (70.6 mIoU on nuScenes), and depth estimation (0.05 REL on KITTI).
We hope that our approach will serve as a solid baseline and facilitate future research.
\end{abstract}
\section{Introduction}
Dense prediction tasks are the foundation of computer vision research, including a wide range of perceptual tasks such as semantic segmentation \cite{Cordts_2016_CVPR,zhou2017scene}, depth estimation \cite{geiger2013vision,silberman2012indoor,song2015sun}, and optical flow \cite{fischer2015flownet,geiger2013vision}.
These tasks require correctly predicting the discrete labels or continuous values for all pixels in the image, which provides detailed contextual understanding and enables various applications.

Numerous methods have rapidly improved the result of perception tasks over a short period of time.
In general terms, these methods can be divided into two paradigms: \textit{discriminative-based} 
 \cite{fu2018deep,zhao2017pyramid,xie2021segformer,cheng2022masked} and \textit{generative-based} 
\cite{xie2017adversarial,hendrik2017universal,isola2017image,li2021semantic,yeh2017semantic}. 
The former approach, which directly learns the mapping between input-output pairs and predicts in a single forward step, has become the current de-facto choice due to its simplicity and efficiency.
Whereas, generative models aim at modeling the underlying distribution of the data, conceptually having a greater capacity to handle challenging tasks.
However, they are often restricted by complex architecture customization as well as various training difficulties \cite{salimans2016improved,karras2018progressive,brock2019large}.
\begin{figure}[t!]
    \centering
    \includegraphics[width=\linewidth]{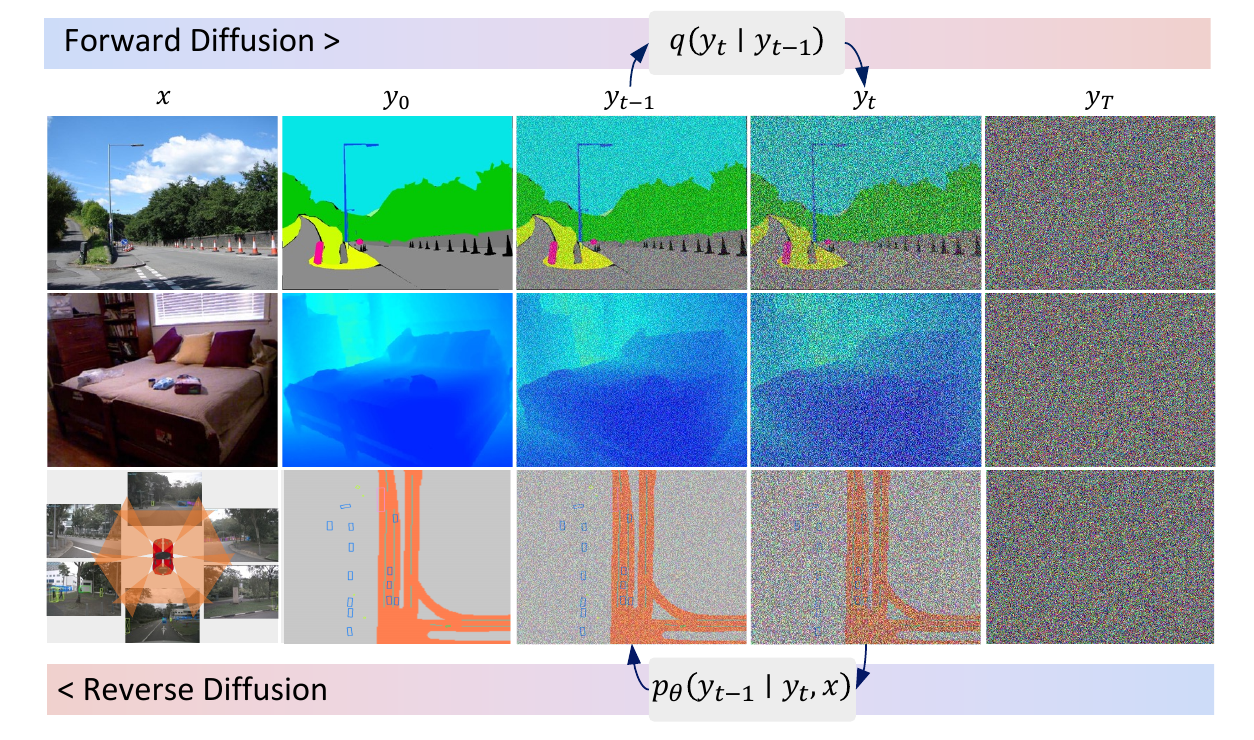}
    \caption{\textbf{Conditional diffusion pipeline for dense visual predictions}. Specifically, a conditional diffusion model is employed, where $q$ is the forward diffusion process and $p_{\theta}$ is the inverse process. The framework iteratively transforms the noise sample $\boldsymbol{y}_T$, drawn from a standard Gaussian distribution, into the desired target prediction $\boldsymbol{y}_0$ under the guidance of the input image $\boldsymbol{x}$.}
    \label{fig_intro_task}
\end{figure}

These challenges have been largely addressed by the diffusion and score-based models~\cite{ho2020denoising,sohl2015deep,song2020improved}.
The solutions, based on \textit{denosing diffusion process}, are conceptually simple:
they apply a continuous diffusion process to transform data into noise and generate new samples by simulating the time-reversed diffusion process.
These methods now enable easy training and achieve superior results on various generative tasks~\cite{nichol2021glide,saharia2022photorealistic,rombach2022high,ramesh2022hierarchical}.
Witnessing these great successes, there has been a recent surge of interest to introduce diffusion models to dense prediction tasks, including semantic segmentation \cite{amit2021segdiff,chen2022generalist,wu2022medsegdiff,wolleb2022diffusion} and depth estimation \cite{saxena2023depthgen}.
However, these methods simply transfer the heavy frameworks from image generation tasks to dense prediction, resulting in low efficiency, slow convergence, and sub-optimal performance.

\begin{figure*}[t!]
    \centering
    \includegraphics[width=\linewidth]{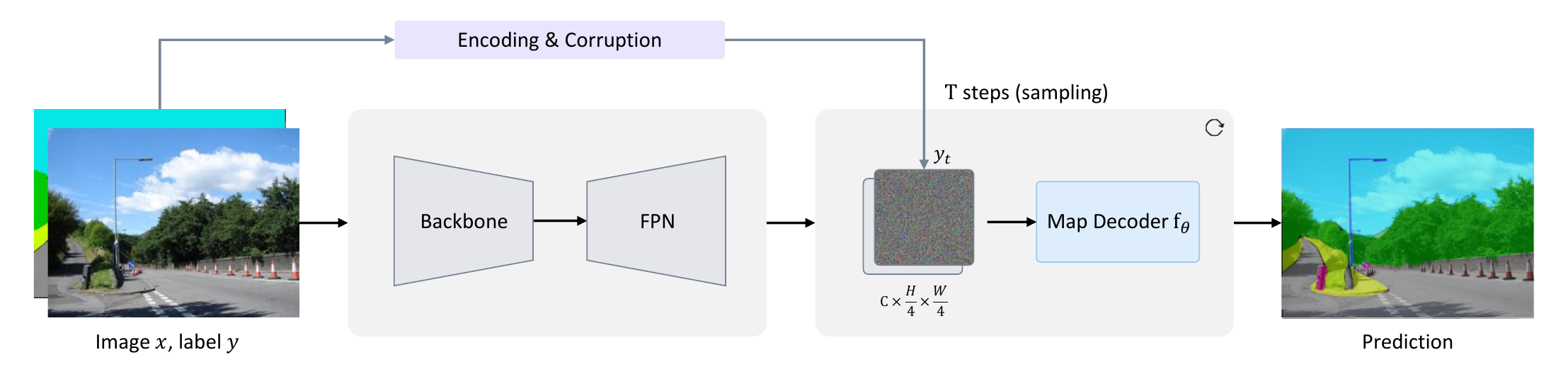}
    \caption{\textbf{The proposed \ours framework}. The image encoder extracts feature representation from the input image $\boldsymbol{x}$ as the condition. The map decoder takes the noisy map $\boldsymbol{y}_t$ as input and produces the denoised prediction under the guidance. During training, the noisy map $\boldsymbol{y}_t$ is constructed by adding Gaussian noise to the encoded ground truth. In inference, the noisy map $\boldsymbol{y}_t$ is randomly sampled from the Gaussian distribution and iteratively refined to obtain the desired prediction $\boldsymbol{y}_0$.
     }
    \label{fig_intro_overview}
\end{figure*}

In this paper, we introduce a general, simple, yet effective diffusion framework for dense visual prediction.
Our method named as DDP, which extends the denoising diffusion process into
the modern perception pipeline effectively (see \cref{fig_intro_overview}).
During training, the Gaussian noise controlled by a noise schedule \cite{nichol2021improved} is added to the encoded ground truth to obtain the noisy maps.
Then these noisy maps are fused with the conditional features from the image encoder, \eg, Swin Transformer \cite{liu2021swin}.
Finally, these fused features are fed to a lightweight map decoder to produce the predictions without noise.
At the inference phase, \ours generates predictions by reversing the learned diffusion process, which adjusts a noisy Gaussian distribution to the learned map distribution under the guidance of the test images (see \cref{fig_intro_task}).

Compared to previous cumbersome diffusion perception models \cite{wu2022medsegdiff,wolleb2022diffusion,saxena2023depthgen}, \ours decouples the image encoder and map decoder. The image encoder runs only once, while the diffusion process is performed only in the lightweight decoder head.
With this efficient design, our proposed method can easily be applied to modern perception tasks.
Furthermore, unlike previous single-step discriminative models, \ours is capable of performing iterative inference multiple times using the shared parameters and exhibits the following appealing properties: (1) dynamic inference to trade off computation and prediction quality and (2) natural awareness of the prediction uncertainty.

We evaluate \ours on three representative dense prediction tasks, including semantic segmentation, BEV map segmentation, and depth estimation, using six popular datasets (ADE20K \cite{zhou2017scene}, Cityscapes \cite{Cordts_2016_CVPR}, nuScenes~\cite{caesar2020nuscenes}, KITTI \cite{geiger2013vision}, NYU-DepthV2 \cite{silberman2012indoor}, and SUN RGB-D \cite{song2015sun}).
Our experimental results demonstrate that DDP significantly outperforms existing state-of-the-art methods. Specifically, on ADE20K, DDP achieves 46.1 mIoU with a single sampling step, which is significantly better than UperNet \cite{xiao2018unified} and K-Net \cite{zhang2021k}. On nuScenes, DDP yields an mIoU of 70.3, which is clearly better than the BEVFusion~\cite{liu2022bevfusion} baseline that achieves an mIoU of 62.7.
Furthermore, by increasing the sampling steps, DDP can achieve even higher performance on both ADE20K and nuScenes, reaching anmIoU of 47.0 and 70.6, respectively. 
Moreover, the gains are more versatile for different model architectures as well as model sizes. \ours achieves 83.9 mIoU on Cityscapes with the ConvNeXt-L backbone and produces a leading REL of 0.05 on KITTI with the Swin-L backbone.

Overall, our contributions in this work are three-fold. 
\begin{itemize}
  \item We formulate the dense visual prediction tasks as a general conditional denoising process, with simple yet highly effective designs.
  \item Our ``noise-to-map" generative paradigm offers several appealing properties, such as the ability to perform dynamic inference and uncertain awareness.
  \item We conduct extensive experiments on three representative tasks with six diverse benchmarks. The results demonstrate that our method, which we refer to as \ours, achieves competitive performance when compared to previous discriminative methods.
\end{itemize}

\section{Related Work}

\paragraph{Diffusion Model.}
Diffusion~\cite{ho2020denoising, sohl2015deep} and score-based generative models~\cite{songdenoising} have been particularly successful as generative models and achieve impressive results across various modalities, including images~\cite{ramesh2022hierarchical,saharia2022image,dhariwal2021diffusion,nichol2021glide,daras2022multiresolution,daras2022multiresolution}, video~\cite{ho2022video,hong2022cogvideo}, audio~\cite{kolesnikov2020big}, and biomedical~\cite{anand2022protein, trippe2022diffusion,schneuing2022structure,corso2022diffdock}.
Given the notable achievements of diffusion models in these respective domains, leveraging such models to develop generation-based perceptual models would prove to be a highly promising avenue to push the boundaries of perceptual tasks to newer heights.

\paragraph{Dense Prediction.}
The perception of real-world scenes via pixel-by-pixel classification or regression is commonly formulated as dense prediction tasks, such as semantic segmentation \cite{Cordts_2016_CVPR,zhou2017scene}, depth estimation \cite{geiger2013vision,silberman2012indoor,song2015sun}, and optical flow \cite{fischer2015flownet,geiger2013vision}.
Numerous methods have emerged and achieved tremendous progress, and these advances can be roughly divided to:
multi-scale feature aggregation \cite{chen2017deeplab, chen2018encoder,xiao2018unified}, high-capacity backbone \cite{xie2021segformer,zheng2021rethinking,ranftl2021vision} and powerful decoder head \cite{strudel2021segmenter,zhang2021k,cheng2021per,jain2022oneformer}.
In this paper, as shown in \cref{fig_intro_task}, which differs from previous discriminative-based methods, we explore a generative ``noise-to-map" paradigm for general dense prediction tasks.

\paragraph{Diffusion Models for Dense Prediction.}
With the recent success of diffusion models in generation tasks, there has been a noticeable rise in interest to incorporate them into dense visual prediction tasks.
Several pioneering works~\cite{wu2022medsegdiff,amit2021segdiff,wolleb2022diffusion,chen2022generalist,saxena2023depthgen,chen2022diffusiondet} attempted to apply the diffusion model to visual perception tasks, \eg image segmentation or depth estimation task. For example, Wolleb \etal \cite{wolleb2022diffusion} explore the diffusion model for medical image segmentation. Pix2Seq-D~\cite{chen2022generalist} applies the bit diffusion model~\cite{chen2023analog} for panoptic segmentation. Our concurrent work DepthGen~\cite{saxena2023depthgen} involves diffusion pipeline to the task of depth estimation.
For all the diffusion models listed above, one or two parameter-heavy convolutional U-Nets~\cite{ronneberger2015u} are adopted, leading to low efficiency, slow convergence, and sub-optimal performance.
In this work, as illustrated in \cref{fig_intro_overview}, we introduce a simple yet effective diffusion framework, which extends the denoising diffusion process into the modern perception pipeline while maintaining accuracy and efficiency.

\section{Methodology}
\subsection{Preliminaries}

\paragraph{Dense Prediction.} The objective of dense prediction tasks is to predict discrete labels or continuous values, denoted as $\boldsymbol{y}$, for every pixel present in the input image $\boldsymbol{x} \in \mathbb{R}^{3 \times h \times w}$.

\paragraph{Conditional Diffusion Model.}
The conditional diffusion model, which is an extension of the diffusion model \cite{ho2020denoising,sohl2015deep,song2020improved}, belongs to the category of likelihood-based models inspired by non-equilibrium thermodynamics.
The conditional diffusion model assumes a forward noising process by gradually adding noise to the data sample, which is defined as:
\begin{equation}
 q\left(\boldsymbol{z}_t \mid \boldsymbol{z}_0\right)= \mathcal{N}\left(\boldsymbol{z}_t ; \sqrt{\bar{\alpha}_t} \boldsymbol{z}_0,\left(1-\bar{\alpha}_t\right) \mathbf{I}\right),
 \label{eq:forward}
\end{equation}
which transforms the data sample $\boldsymbol{z}_0$ to a latent noisy sample $\boldsymbol{z}_t$ for $t \in\{0,1, \ldots, T\}$. The constants $\bar{\alpha}_t:=$ $\prod_{s=0}^t \alpha_s=\prod_{s=0}^t\left(1-\beta_s\right)$ and $\beta_s$ represents the noise schedule~\cite{nichol2021improved,ho2020denoising}.
During training, the reverse process model $f_\theta \left(\boldsymbol{z}_t, \boldsymbol{x}, t\right)$ is trained to predict $\boldsymbol{z}_0$ from $\boldsymbol{z}_t$ under the guidance of condition $\boldsymbol{x}$ by minimizing the training objective function (\ie, $l_2$ loss).
At the inference stage, predicted data sample $\boldsymbol{z}_0$ is reconstructed from a random noise $\boldsymbol{z}_T$ with the model $f_\theta$, conditional input $\boldsymbol{x}$, and a translation rule~\cite{ho2020denoising,song2020denoising} in a markovian way, \ie, $\boldsymbol{z}_T \rightarrow \boldsymbol{z}_{T-\Delta} \rightarrow \ldots \rightarrow \boldsymbol{z}_0$, which can be formulated as:
\begin{equation}
p_\theta\left(\boldsymbol{z}_{0: T} \mid \boldsymbol{x}\right)=p\left(\boldsymbol{z}_T\right) \prod_{t=1}^T p_\theta\left(\boldsymbol{z}_{t-1} \mid \boldsymbol{z}_t, \boldsymbol{x}\right).  
\end{equation}
In this paper, our goal is to solve dense prediction tasks via the conditional diffusion model.
In our setting, the data samples are the ground truth map $\boldsymbol{z}_0=\boldsymbol{y}$, and a neural network $f_\theta$ is trained to predict $\boldsymbol{z}_0$ from random noise $\boldsymbol{z}_t \sim \mathcal{N}(0, \mathbf{I})$ conditioned on the corresponding image $\boldsymbol{x}$.

\subsection{Architecture}
Since the diffusion model generates samples progressively, it requires multiple runs of the model in the inference stage.
Previous methods~\cite{wu2022medsegdiff,saxena2023depthgen, wolleb2022diffusion} apply the model $f_\theta$ in multiple steps on the raw image $\boldsymbol{x}$, which significantly increases the computational overhead.
To alleviate this issue, we separate the entire model into two parts: image encoder and map decoder, as shown in \cref{fig_intro_overview}.
The image encoder forwards only once to extract the feature map from the input image $\boldsymbol{x}$. Then the map decoder employs it as the condition rather than the raw image $\boldsymbol{x}$, to gradually refine the prediction from the noisy map $\boldsymbol{y}_t$.
\paragraph{Image Encoder.}
The image encoder receives the raw image $\boldsymbol{x}$ as input and generates multi-scale features at 4 different resolutions.
Subsequently, these multi-scale features are fused using the FPN \cite{lin2017feature} and aggregated by a 1$\times$1 convolution.
The produced feature map, with the resolution of $256\times \frac{h}{4} \times \frac{w}{4}$, is employed as the condition for the map decoder.
In contrast to the previous methods~\cite{amit2021segdiff, wu2022medsegdiff, saxena2023depthgen}, \ours is able to work with modern network architectures such as ConvNext \cite{liu2022convnet} and Swin Transformer \cite{liu2021swin}.

\paragraph{Map Decoder.}
The map decoder $f_\theta$ takes as input the noisy map $\boldsymbol{y}_t$ and the feature map from the image encoder via concatenation and performs a pixel-by-pixel classification or regression.
Following the common practice \cite{cheng2022masked,zhu2020deformable,zhang2022dino} in modern perception pipelines, we simply stack six layers of deformable attention as the map decoder.
Compared to previous works \cite{amit2021segdiff, wu2022medsegdiff, saxena2023depthgen, chen2022generalist, wolleb2022diffusion} that use the parameter-intensive U-Nets, our map decoder is lightweight and compact, allowing efficient reuse of the shared parameters during the multi-step reverse diffusion process.

\begin{algorithm}[t!]
\caption{\ours Training}
\label{algo:ddp:training}
\definecolor{codeblue}{HTML}{2E8B57} 
\definecolor{codekw}{HTML}{DC143C} 
\lstset{
  backgroundcolor=\color{white},
  columns=fullflexible,
  breaklines=true,
  captionpos=b,
  commentstyle=\fontsize{7.2pt}{7.2pt}\color{codeblue},
  keywordstyle=\fontsize{7.2pt}{7.2pt}\color{codekw},
  escapechar={|}, 
}
\lstset{language=Python}
\begin{lstlisting}[xleftmargin=-1em]
def train(images, maps):
  """images: [b, 3, h, w], maps: [b, 1, h, w]"""
  img_enc = image_encoder(images) # encode image
  map_enc = encoding(maps) # encode gt
  map_enc = (sigmoid(map_enc) * 2 - 1) * scale
  # corrupt gt
  t, eps = uniform(0, 1), normal(mean=0, std=1)
  map_crpt = sqrt(alpha_cumprod(t)) * map_enc +
              sqrt(1 - alpha_cumprod(t)) * eps
  # predict and backward
  map_pred = map_decoder(map_crpt, img_enc, t)
  loss = objective_func(map_pred, maps)
  return loss
\end{lstlisting}
\end{algorithm}

\subsection{Training}
During training, we first construct a diffusion process from the ground truth $\boldsymbol{y}$ to the noisy map $\boldsymbol{y}_t$ and then train the model to reverse this process.
The training procedure for \ours is provided in \cref{algo:ddp:training} (for more details please refer to \cref{supp:diffusion}).

\paragraph{Label Encoding.}
\label{para:label_encoding}
Standard diffusion models assume continuous data, which makes them a convenient choice for regression tasks with continuous values (\eg, depth estimation). However, existing studies~\cite{chen2022generalist,chen2023analog} show that they are unsuitable for discrete labels (\eg, semantic segmentation).
Therefore, we explore several encoding strategies for the discrete labels, including:
(1) One-hot encoding, which represents categorical labels as binary vectors of 0 and 1;
(2) Analog bits encoding~\cite{chen2022generalist}, which first converts discrete integers into bit strings, and then casts them as real numbers;
(3) Class embedding, which uses a learnable embedding layer to project discrete labels into a high-dimensional continuous space, with a sigmoid function for normalization.
For all of these strategies, we normalize and scale the range of encoded labels within $[-{\rm scale}, +{\rm scale}]$, as shown in \cref{algo:ddp:training}.
Notably, the scaling factor $\rm scale$ controls the signal-to-noise ratio (SNR) \cite{chen2022generalist,chen2023importance}, which is an important hyper-parameter for diffusion models.
We compare these strategies in \cref{tab:label_encoding} and find class embedding work best.
More discussions are in \cref{sec:exp:abla}.

\paragraph{Map Corruption.}
We add Gaussian noise to corrupt the encoded ground truth, obtaining the noisy map $\boldsymbol{y}_t$.
As shown in \cref{eq:forward}, the intensity of corruption noise is controlled by $\alpha_t$, which adopts the monotonically decreasing schedule for $\alpha_t$ in different time steps $t \in [0,1]$.
Different noise scheduling strategies, including cosine schedule \cite{nichol2021improved} and linear schedule \cite{ho2020denoising},
are compared and discussed in \cref{sec:exp:abla}.
We found that the cosine schedule usually worked best in our benchmark tasks.

\paragraph{Objective Function.}
Standard diffusion models are trained with $l_2$ loss, which is reasonable for dense prediction tasks, but we found that adopting a task-specific loss works better for supervision, \eg, cross-entropy loss for semantic segmentation, sigloss for depth estimation.

\begin{algorithm}[t!]
\caption{\ours Sampling}
\label{algo:ddp:sampling}
\definecolor{codeblue}{HTML}{2E8B57}
\definecolor{codekw}{HTML}{DC143C}
\lstset{
  backgroundcolor=\color{white},
  columns=fullflexible,
  breaklines=true,
  captionpos=b,
  commentstyle=\fontsize{7.5pt}{7.5pt}\color{codeblue},
  keywordstyle=\fontsize{7.5pt}{7.5pt}\color{codekw},
  escapechar={|}, 
}
\lstset{language=Python}
\begin{lstlisting}[xleftmargin=-1em]
def sample(images, steps, td=1):
  """steps: sample steps, td: time difference"""
  img_enc = image_encoder(images)
  map_t = normal(0, 1) # [b, 256, h/4, w/4]
  for step in range(steps):
    # time intervals
    t_now = 1 - step / steps
    t_next = max(1 - (step + 1 + td) / steps, 0)
    # predict map_0 from map_t
    map_pred = map_decoder(map_t, img_enc, t_now)
    # estimate map_t at t_next
    map_t = ddim(map_t, map_pred, t_now, t_next)
  return map_pred
\end{lstlisting}
\end{algorithm}

\subsection{Inference}
\label{sec:inference}
Given a test image as condition input, the model starts with a random noise map sampled from a Gaussian distribution and gradually refines the prediction, we summarize the inference procedure in \cref{algo:ddp:sampling}.

\paragraph{Sampling Rule.}
We choose the DDIM update rule~\cite{song2020denoising} for the sampling.
In each sampling step $t$, the random noise $\boldsymbol{y}_T$ or the predicted noisy map $\boldsymbol{y}_{t+1}$ from the last step is fused with the conditional feature map, and sent to the map decoder $f_\theta$ for map prediction.
After getting the predicted result of the current step, we compute the noisy map $\boldsymbol{y}_{t}$ for the next step using the reparameterization trick.
Following~\cite{chen2022analog,chen2022generalist,chen2022diffusiondet}, we use the asymmetric time intervals (controlled by a hyper-parameter $td$) during the inference stage, and $td=1$ works best in our method.

\paragraph{Sampling Drift.}
As displayed in \cref{fig:multiple_inference}, we empirically observe that the model performance improves in a few sampling steps and then declines slightly as the number of steps increases.
Similar observations can also be found in \cite{chen2022diffusiondet,chen2022sampling,saxena2023depthgen}.
This performance decline can be attributed to the ``sampling drift'' challenge, which refers to the discrepancy between the distribution of training and sampling data.
During training, the model is trained to inverse the noisy ground truth map, while during testing, the model is inferred to remove noise from its ``imperfect" prediction, which drifts away from the underlying corrupted distributions.
This drift becomes pronounced with smaller time steps $t$,  owing to the compounded errors, and is further intensified when a sample deviates more substantially from the distribution of ground truth~\cite{daras2023consistent}.
%

To verify our hypothesis, in the last 5k iterations of training, we construct $\boldsymbol{y}_t$ using the model's prediction rather than the ground truth.
The approach transforms the training target to remove the added noise on its own predictions, thereby aligning the data distribution of training and testing. 
We name this approach ``\emph{self-aligned denoising}."
As revealed in \cref{fig:multiple_inference}, this approach tends to produce saturation instead of performance degradation. 
Our findings suggest that incorporating the diffusion process into perception tasks could enhance efficacy compared to image generation (\eg, about 50 DDIM steps for image generation).
In other words, the proposed DDP can improve efficiency (\eg, satisfied results in 3 iterative steps) while retaining the benefits of the diffusion model.
More discussions can be found in \cref{supp:diffusion}.

\paragraph{Multiple Inference.}
By virtue of the multi-step sampling procedure, our method supports dynamic inference, which has the flexibility to trade compute for prediction quality.
Besides, it naturally enables the assessment of the reliability and uncertainty of model predictions.

\section{Experiment}
We first present the appealing properties of our \ours, followed by empirical evaluations of its performance against leading methods on several representative  tasks, including semantic segmentation, BEV map segmentation, and monocular depth estimation. 
Finally, we provide ablation studies on the \ours components. 
Due to space limitations, 
more implementation details and experimental results are provided in \cref{supp:implement} and \cref{supp:experiment}, respectively.

\subsection{Main Properties}
\begin{figure}
  \centering
  \subfloat[\textbf{Dynamic inference}.~The results of multiple inference on Cityscapes.]{\includegraphics[width=0.95\linewidth]{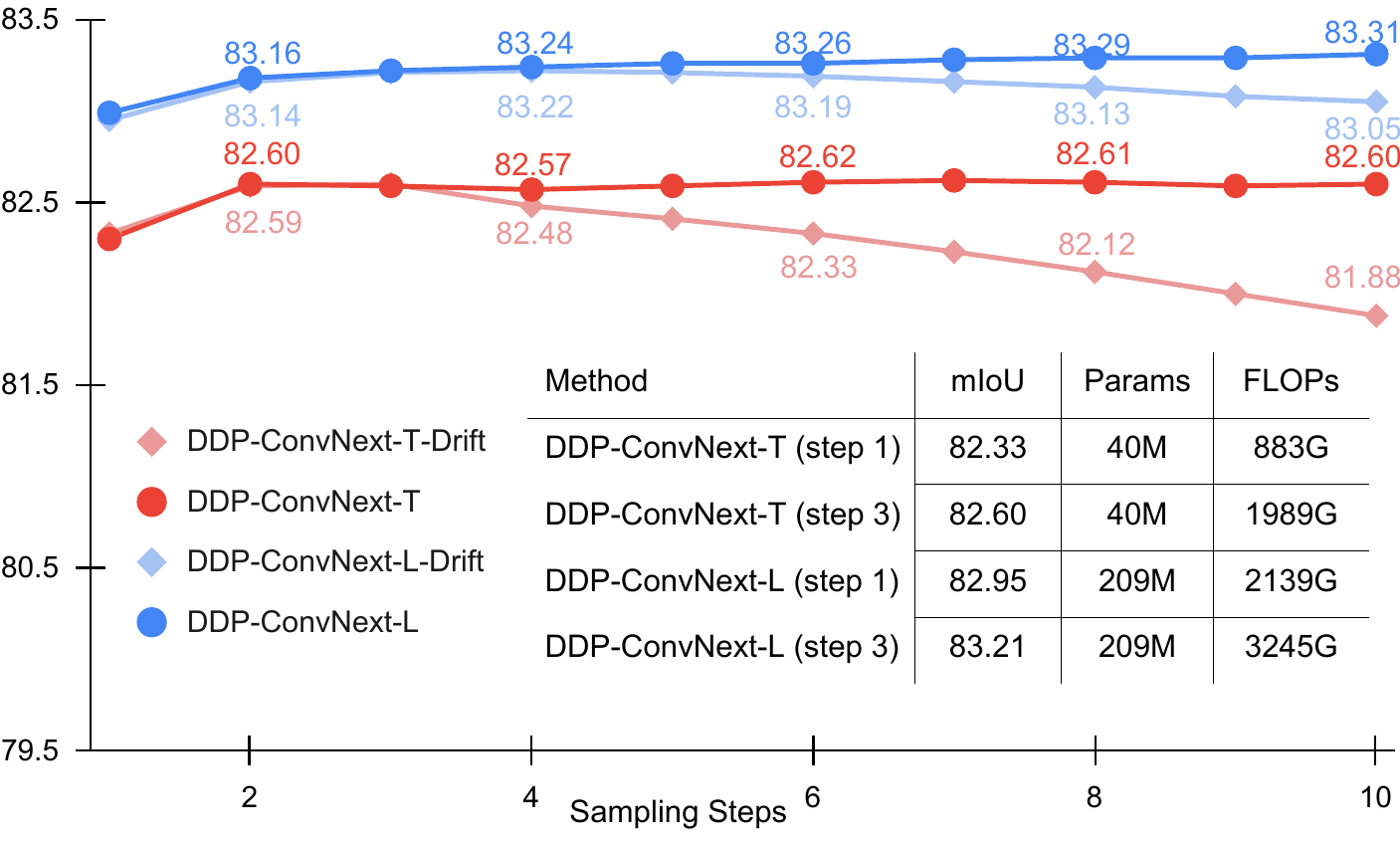}
  \label{fig:multiple_inference}}
  \vspace{-0.5em}
  \\
  \subfloat[\textbf{Inference trajectory}.~Predicted mask results on different time steps.]{\includegraphics[width=0.95\linewidth]{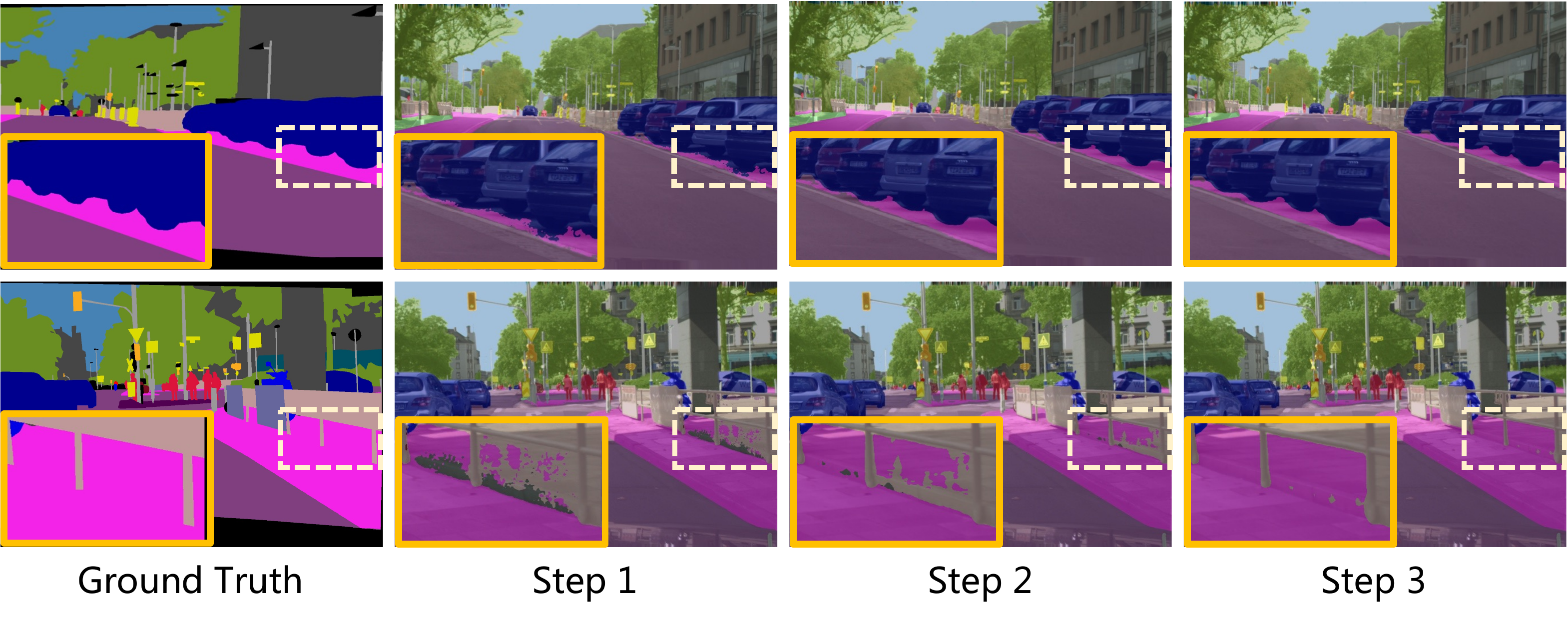}
  \label{fig:refine}}
  \vspace{-0.5em}
  \\
  \subfloat[\textbf{Uncertainty awareness}. High response areas in the uncertainty map indicate high estimated uncertainty and are highly positively correlated with white areas in the error map, which indicate misclassified points. Zoom in for better visualization.]{\includegraphics[width=0.95\linewidth]{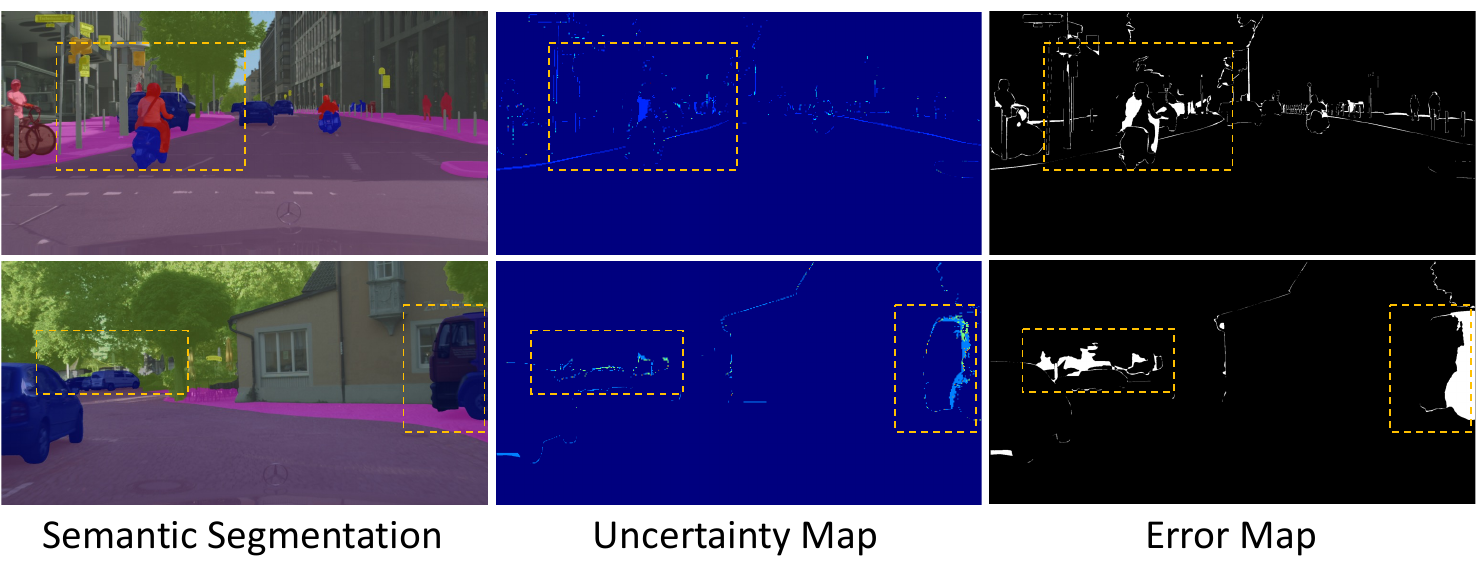}
  \label{fig:uncertainty}}
  \vspace{1em}
  \caption{\textbf{DDP enjoys two appealing properties:} dynamic inference to trading off computation and prediction quality and the natural awareness of the prediction uncertainty.}
  \label{fig:properties}
\end{figure}

We explore and show properties of \ours in \cref{fig:properties} using the default setting in \cref{sec:semantic_settings}.
With such a multi-step sampling procedure, we have the flexibility to trade computational cost for prediction quality.
Furthermore, the stochastic sampling process allows the computing of pixel-wise uncertainty maps of the prediction.

\paragraph{Dynamic Inference.}
We evaluate \ours with ConvNext-T and ConvNext-L backbones by increasing their sampling steps from 1 to 10.
The results are presented in \cref{fig:multiple_inference}.
It can be seen that the DDP can continuously improve its performance by using more sampling steps.
For example, DDP with ConvNext-T shows an increase from 82.33 mIoU (1 step) to 82.60 mIoU (3 steps), and we visualize the inference trajectory in \cref{fig:refine}.
In comparison to the previous single-step method, our approach boasts the flexibility to balance computational cost against accuracy. This means our method can be adapted to different trade-offs between speed and accuracy under various scenarios without the need to retrain the network.

\paragraph{Uncertainty Awareness.}
In addition to the performance gains, the proposed \ours can naturally provide uncertainty estimates.
In the multi-step sampling process, we can simply count the pixels where the predicted result of each step differs from the result of the previous step, and finally, we simply normalize this change count map to 0-1 and obtain an uncertainty map.
In comparison, DDP is naturally and easily capable of estimating uncertainty, whereas previous methods \cite{loquercio2020general,harakeh2020bayesod} require complicated modeling such as Bayesian networks.

\subsection{Semantic Segmentation}

\paragraph{Datasets.}
We evaluate the proposed \ours using two widely used datasets: ADE20K \cite{zhou2017scene} and Cityscapes \cite{Cordts_2016_CVPR}.
ADE20K is a large-scale scene parsing dataset with over 20,000 images, and Cityscapes is a street scene dataset with high-quality pixel-level annotations for 5,000 images.

\paragraph{Settings.}
\label{sec:semantic_settings}
In the training phase, following common practices \cite{wang2021pyramid,chen2022vitadapter,xie2021segformer,wang2022internimage}, the crop size is set to 512$\times$512 for ADE20K, and 512$\times$1024 for Cityscapes.
We optimize our \ours models using the AdamW~\cite{loshchilov2017decoupled} optimizer, with an initial learning rate of $6\times10^{-5}$ and a weight decay of 0.01.
All models are trained for 160k iterations and compared fairly with previous non-diffusion methods.

\begin{table}[t]
\centering
\footnotesize
\renewcommand\arraystretch{1.05}
\setlength{\tabcolsep}{1.35mm}
\begin{tabular}{l|l|c|c|c|c}
    Method & Backbone  & \#Param & FLOPs  & mIoU & +MS \\
    \hline
    UperNet~\cite{xiao2018unified} & Swin-T & 60M & 236G & 44.5 & 45.8\\
    Region Rebalance~\cite{cui2022region} & Swin-T & 60M & 236G & 45.0 & 46.5\\
    MaskFormer~\cite{cheng2021per}  & Swin-T & 42M & 55G & 46.7 & 48.8 \\
    Mask2Former~\cite{cheng2022masked} & Swin-T & 47M & 74G & 47.7 & 49.6 \\    
    K-Net~\cite{zhang2021k} & Swin-T & 73M & 256G & 45.8 & 46.3\\
    SenFormer~\cite{bousselham2021efficient} & Swin-T & 144M & 179G & 46.0 & 46.4\\
    \rowcolor{gray!10} 
    Non-diffusion Baseline & Swin-T & 35M & 111G & 44.9 & 46.1 \\  
    \rowcolor{gray!10} 
    \ours (step 1) & Swin-T & 40M & 113G & 46.1 & 47.6 \\
    \rowcolor{gray!10} 
    \ours (step 3) & Swin-T & 40M & 252G & 47.0 & 47.8\\
    \hline
    UperNet~\cite{xiao2018unified} & Swin-S & 81M & 259G & 47.6 & 49.5\\ 
    \rowcolor{gray!10} 
    \ours (step 1) & Swin-S & 61M & 136G & 48.4 & 49.7 \\
    \rowcolor{gray!10} 
    \ours (step 3) & Swin-S & 61M & 276G & 48.7 & 49.7 \\
    \hline
    UperNet~\cite{xiao2018unified} & Swin-B & 121M & 297G & 48.1 & 49.7\\
    \rowcolor{gray!10} 
    \ours (step 1) & Swin-B & 99M & 173G & 49.2 & 50.8\\
    \rowcolor{gray!10} 
    \ours (step 3) & Swin-B & 99M & 312G & 49.4 & 50.8 \\
    \hline
    UperNet~\cite{xiao2018unified} & Swin-L$^\dagger$ & 234M & 411G & 52.1 & 53.5\\
    \rowcolor{gray!10} 
    \ours (step 1) & Swin-L$^\dagger$ & 207M & 285G & 53.1 & 54.4 \\
    \rowcolor{gray!10} 
    \ours (step 3) & Swin-L$^\dagger$ & 207M & 425G & 53.2 & 54.4 \\
    \end{tabular}
    \vspace{1em}
    \caption{\textbf{Semantic segmentation on ADE20K val set.}
    We report single-scale (SS) and multi-scale (MS) mIoU.
    The FLOPs are measured with 512$\times$512 inputs.
    Backbones pre-trained on ImageNet-22K are marked with $^\dagger$.
    }
    \label{tab:result_ade20k}
\end{table}

\paragraph{Results on ADE20K.}
Table \ref{tab:result_ade20k} presents the semantic segmentation performance of \ours on  ADE20K \cite{zhou2017scene}, which shows that our method consistently outperforms many representative methods \cite{xiao2018unified,cui2022region,zhang2021k,bousselham2021efficient} and the non-diffusion baseline across different backbones.
For instance, when using Swin-T \cite{liu2021swin} as the backbone, our \ours (step 1) yields a promising result of 46.1 mIoU, surpassing the non-diffusion baseline (\ours \emph{w/o} diffusion process)  by 1.2 points (46.1 \emph{vs.}~44.9).
Moreover, our \ours (step 3) can further enhance the performance to 47.0 mIoU, attaining a remarkable gain of 0.9 points by multi-steps of denoising diffusion.
With the Swin-L backbone, our \ours (step 3) achieves the best performance of 53.2 mIoU, which is 1.1 points (53.2 \emph{vs.}~52.1) better than UperNet with comparable FLOPs.
These results suggest that our \ours not only achieves a performance gain but also offers more flexibility than previous methods.

\begin{table}[t!]
\centering
\footnotesize
\renewcommand\arraystretch{1.05}
\setlength{\tabcolsep}{1.25mm}
    \begin{tabular}{l|l|c|c|c|c}
        Method & Backbone & \#Param  & FLOPs & mIoU & +MS \\
        \hline
        Segmenter~\cite{strudel2021segmenter} & ViT-L$^\dagger$ & 333M & 2685G & 79.10 & 81.30\\
        SETR-PUP~\cite{zheng2021rethinking} & ViT-L$^\dagger$  & 318M & 2955G &  79.34 & 82.15\\
        StructToken~\cite{lin2022structtoken} & ViT-L$^\dagger$  & 364M  & 2913G & 80.05 & 82.07 \\
        OCRNet~\cite{yuan2020object,yuan2021hrformer} & HRFormer-B & 56M & 2240G & 81.90 & 82.60 \\
        SegFormer-B5~\cite{xie2021segformer} & MiT-B5 & 85M  & 1448G & 82.25 & 83.48  \\
        DiversePatch~\cite{gong2021vision} & Swin-L$^\dagger$ & 234M & 3190G  & 82.70 & 83.60 \\ 
        Mask2Former~\cite{cheng2022masked} & Swin-L$^\dagger$ & 216M   & 2113G & 83.30 & 84.30 \\
        
        \hline
        \ours (step 1) & Swin-T & 39M &885G  & 80.96 & 82.25 \\
        \ours (step 3) & Swin-T & 39M &1992G & 81.24 & 82.46 \\
        \ours (step 1) & Swin-S & 61M &1067G & 82.17 & 83.06 \\
        \ours (step 3) & Swin-S & 61M &2174G & 82.41 & 83.21 \\
        \rowcolor{gray!10} 
        \ours (step 1) & Swin-B & 99M &1357G & 82.37 & 83.36 \\
        \rowcolor{gray!10} 
        \ours (step 3) & Swin-B & 99M &2464G & 82.54 & 83.42    \\   
        \hline
        \ours (step 1) & ConvNext-T  & 40M  & 883G  &82.33 & 83.00 \\
        \ours (step 3) & ConvNext-T  & 40M  & 1989G &82.60 & 83.15 \\
        \ours (step 1) & ConvNext-S  & 62M  & 1059G &82.37 & 83.38 \\
        \ours (step 3) & ConvNext-S  & 62M  & 2166G &82.69 & 83.58 \\
        \ours (step 1) & ConvNext-B  & 100M & 1340G &82.59 & 83.47 \\
        \ours (step 3) & ConvNext-B  & 100M & 2447G &82.78 & 83.49 \\
        \rowcolor{gray!10} 
        \ours (step 1) & ConvNext-L$^\dagger$ & 209M & 2139G &82.95 & 83.76 \\
        \rowcolor{gray!10} 
        \ours (step 3) & ConvNext-L$^\dagger$ & 209M & 3245G &
        83.21 & 83.92\\
    \end{tabular}
    \vspace{1em}
\caption{\textbf{Semantic segmentation on Cityscapes val set.}
We report single-scale (SS) and multi-scale (MS) mIoU.
The FLOPs are measured with 1024$\times$2048 inputs.
Backbones pre-trained on ImageNet-22K are marked with $^\dagger$.
}
\label{tab:result_cityscapes}
\end{table}

\paragraph{Results on Cityscapes.}
We compare our \ours with various representative models on Cityscapes~\cite{Cordts_2016_CVPR} in Table \ref{tab:result_cityscapes}, such as Segmenter \cite{strudel2021segmenter}, SETR \cite{zheng2021rethinking}, SegFormer \cite{xie2021segformer}, DiversePatch \cite{gong2021vision}, and Mask2Former \cite{cheng2022masked}, and so on.
As shown, we conduct extensive experiments based on ConvNeXt \cite{liu2022convnet} and Swin \cite{liu2021swin} with different model sizes. 
When using ConvNeXt-L$^\dagger$ as the backbone, our \ours (step 1) produces a competitive result of 82.95 mIoU, and it can be further boosted to 83.21 mIoU (step 3). 
This phenomenon was also observed when taking Swin-T as the backbone, and the mIoU increased from 80.96 to 81.24 through additional 2 sampling steps.
These experimental results demonstrate the scalability of our methodology, which can be applied to different model structures of arbitrary size.
Moreover, once again, the experimental results show that \ours achieves progressive improvements through multi-step denoising diffusion while keeping comparable computational overhead.

\paragraph{Discussion.}  
The original intention of DDP is to design a diffusion-based general framework for various dense prediction tasks. 
Although its segmentation performance is slightly lower than its specialized counterpart Mask2Former \cite{cheng2022masked}, 
it remains highly competitive and has several attractive features.
How to design a segmentation-specific diffusion framework to achieve better performance than Mask2Former is left for future research.

\subsection{BEV Map Segmentation}

\paragraph{Dataset.} 
We conduct our experiments of BEV map segmentation on the nuScenes~\cite{caesar2020nuscenes} dataset. 
It is a large-scale autonomous driving perception dataset, which includes over 1000 urban road scenes covering different time periods and weather conditions in two cities, Boston and Singapore.

\paragraph{Settings.}
We further verify the \ours framework on the BEV map segmentation task.
Specifically, we equip our method with the representative method BEVFusion \cite{liu2022bevfusion}, where we directly replace its segmentation head with the proposed map decoder for the diffusion process.
We follow evaluation protocol from~\cite{liu2022bevfusion} and compare the results with state-of-the-art methods~\cite{xie2022m2bev, yin2021multimodal, liu2022bevfusion, borse2023x}.
We report the IoU of 6 background classes, including drivable space (Dri), pedestrian crossing (Ped), walk-way (Wal), stop line (Sto), car-parking area (Car), and lane divider (Div), and use the mean IoU as the primary evaluation metric.
Other training settings are kept the same as \cite{liu2022bevfusion} for fair comparisons.

\paragraph{Results.} 
We show the results of our BEV map segmentation experiments in Table \ref{tab:exp:bev:nuscenes}, which exhibit the superior performance of our approach, over existing state-of-the-art methods.
Specifically, in the camera-only scenario, our \ours (step 1) attains a 59.3 mIoU score on the nuScenes validation dataset, which surpasses the previous best method X-Align \cite{borse2023x} by 1.3 mIoU (59.3 \emph{vs.} 58.0). 
By iteratively refining the output of the model, \ours (step 3) sets a new state-of-the-art record of 59.4 mIoU solely based on camera modality.
In the multi-modality setting, we improve the segmentation results of our \ours (step 1) to 70.3 mIoU by combining LiDAR information, significantly higher than the current state-of-the-art methods \cite{liu2022bevfusion, borse2023x} by at least 4.6 mIoU. Remarkably, this performance can be further enhanced to a maximum of 70.6 mIoU by leveraging the benefits of iterative denoising diffusion.
In summary, these results demonstrate that \ours can be easily generalized to other tasks and obtain performance gains, proving the effectiveness and generalization of our approach.

\begin{table}[t]
    \centering
    \footnotesize
    \renewcommand\arraystretch{1.05}
    \setlength{\tabcolsep}{0.88mm}
    \begin{tabular}{l|c|cccccc|c}
        Method & Modality & Dri & Ped & Wal & Sto & Car & Div & Mean \\
        \hline 
        OFT~\cite{roddick2018orthographic} & C & 74.0 & 35.3 & 45.9 & 27.5 & 35.9 & 33.9 & 42.1 \\
        LSS~\cite{philion2020lift}  & C & 75.4 & 38.8 & 46.3 & 30.3 & 39.1 & 36.5 & 44.4 \\
        CVT~\cite{zhou2022cross}  & C &  74.3 & 36.8 & 39.9 & 25.8 & 35.0 & 29.4 & 40.2 \\
        M$^2$BEV~\cite{xie2022m2bev} & C & 77.2 & - & - & - & - & 40.5 & - \\
        BEVFusion~\cite{liu2022bevfusion} & C & 81.7 & 54.8 & 58.4 & 47.4 & 50.7 & 46.4 & 56.6 \\
        X-Align~\cite{borse2023x} & C & 82.4 & 55.6 & 59.3 & 49.6 & 53.8 & 47.4 & 58.0 \\ 
        \rowcolor{gray!10} 
        \ours (step 1) & C &83.2  &58.5 &61.6 &52.4 &51.1 &48.9 &59.3 \\
        \rowcolor{gray!10} 
        \ours (step 3) & C & 83.6 &58.3 &61.8 &52.3 &51.4 &49.2 &\textbf{59.4} \\
        \hline 
        PointPainting~\cite{vora2020pointpainting}  & C+L & 75.9 & 48.5 & 57.1 & 36.9 & 34.5 & 41.9 & 49.1 \\
        MVP~\cite{yin2021multimodal} & C+L & 76.1 & 48.7 & 57.0 & 36.9 & 33.0 & 42.2 & 49.0 \\
        BEVFusion~\cite{liu2022bevfusion} & C+L & 85.5 & 60.5 & 67.6 & 52.0 & 57.0 & 53.7 & 62.7 \\
        X-Align~\cite{borse2023x} & C+L & 86.8 & 65.2 & 70.0 & 58.3 & 57.1 & 58.2 & 65.7 \\
        \rowcolor{gray!10} 
        \ours (step 1)  & C+L & 89.3   &69.5   &74.8  &62.5 &63.5   & 62.3    &70.3   \\
        \rowcolor{gray!10} 
        \ours (step 3)  & C+L &89.4   &69.8 &75.0   &63.0   & 63.8   &62.6   &\textbf{70.6} \\
    \end{tabular}
    \vspace{1em}
    \caption{\textbf{BEV map segmentation on nuScenes val set.} 
    We report the IoU of 6 background classes and the mean IoU.
    ``C" and ``L" denotes the camera modality and LiDAR modality, respectively.
    }
    \label{tab:exp:bev:nuscenes}
\end{table}

\subsection{Depth Estimation}
\paragraph{Datasets.}
We evaluate the depth estimation performance of \ours on three prominent datasets, namely KITTI \cite{geiger2013vision}, NYU-DepthV2 \cite{silberman2012indoor}, and SUN RGB-D \cite{song2015sun}.
(1) The KITTI dataset encompasses stereo image pairs and corresponding ground truth depth maps for outdoor scenes captured by a car-mounted camera.
Following common practices \cite{eigen2014depth,li2022depthformer}, we use about 26K left-view images for training and 697 images for testing.
(2) The NYU dataset contains RGB-Depth images for indoor scenes captured at a resolution of 640$\times$480.
Similar to prior research \cite{li2022depthformer}, the model is trained on 24K train images and evaluated on the reserved 652 images.
(3) The SUN RGB-D dataset is a vast collection of around 10K indoor images. 
We employ it to evaluate the generalization abilities of our NYU pre-trained models.
The results on KITTI are shown in the main paper, while others will be provided in the supplementary material.

\paragraph{Settings.}
We incorporate the \ours model into the codebase developed by \cite{li2022depthformer} for depth estimation experiments.
We excluded the discrete label encoding module as the task requires continuous value regression.
All experimental settings are the same as \cite{li2022depthformer} for a fair comparison.

\paragraph{Metrics.}
Typically, the evaluation of depth estimation methods employs the following metrics: accuracy under threshold ($\delta_i<1.25^i, i = 1, 2, 3$), mean absolute relative error (REL), mean squared relative error (SqRel), root mean squared error (RMSE), root mean squared log error (RMSE log), and mean log10 error (log10).

\begin{table*}[ht]
\centering
\footnotesize
\renewcommand\arraystretch{1.05}
\setlength{\tabcolsep}{2.2mm}
\begin{tabular}{l|l|ccc|cccc}
Method & Backbone & $\delta_1 \uparrow$ & $\delta_2 \uparrow$ & $\delta_3 \uparrow$ & REL $\downarrow$ & SqRel $\downarrow$ & RMSE $\downarrow$ & RMSE $\log \downarrow$ \\
\hline
DORN~\cite{fu2018deep}  & ResNet-101 & 0.932 & 0.984 & 0.994 & 0.072 & 0.307 & 2.727 & 0.120 \\
VNL~\cite{yin2019enforcing} & ResNeXt-101 & 0.938 & 0.990 & \underline{0.998} & 0.072 & - & 3.258 & 0.117 \\
BTS~\cite{lee2019big}  & DenseNet-161 & 0.956 & 0.993 & \underline{0.998} & 0.059 & 0.245 & 2.756 & 0.096 \\
TransDepth~\cite{yang2021transformer}  & ResNet-50 + ViT-B & 0.956 & 0.994 & \textbf{0.999} & 0.064 & 0.252 & 2.755 & 0.098 \\
DPT~\cite{ranftl2021vision}  & ResNet-50 + ViT-B & 0.959 & 0.995 & \textbf{0.999} & 0.062 & - & 2.573 & 0.092 \\
AdaBins~\cite{bhat2021adabins}  & EfficientNet-B5 + Mini-ViT & 0.964 & 0.995 & \textbf{0.999} & 0.058 & 0.190 & 2.360 & 0.088 \\
DepthFormer~\cite{li2022depthformer} & ResNet-50 + Swin-T & 0.966 & 0.995 & \textbf{0.999} & 0.056 & 0.177 & 2.252 & 0.086 \\
DepthFormer~\cite{li2022depthformer} & ResNet-50 + Swin-L$^\dagger$  & \textbf{0.975} & \textbf{0.997} & \textbf{0.999} & 0.052 & 0.158 & 2.143 & 0.079 \\
BinsFormer~\cite{li2022binsformer} & Swin-L$^\dagger$  & \underline{0.974} & \textbf{0.997} & \textbf{0.999} & 0.052 & \underline{0.151} & \underline{2.098} & 0.079 \\
\hline
DepthGen (step 8)*~\cite{saxena2023depthgen} &  Efficient U-Net & 0.953 & 0.991 & \underline{0.998} & 0.064 & 0.356 & 2.985 & 0.100\\
\rowcolor{gray!10} 
\ours (step 3) & Swin-T &0.969 & \underline{0.996} & \textbf{0.999} & 0.054 & 0.168&2.172&0.083 \\ 
\rowcolor{gray!10} 
\ours (step 3) & Swin-S &0.970 & \underline{0.996} & \textbf{0.999} & 0.053 & 0.167 &2.171 & 0.082 \\
\rowcolor{gray!10} 
\ours (step 3) & Swin-B$^\dagger$ & 0.973 & \textbf{0.997} & \textbf{0.999} & \underline{0.051} & 0.155 & 2.119 & \underline{0.078} \\
\rowcolor{gray!20} 
\ours (step 3) & Swin-L$^\dagger$ & \textbf{0.975} & \textbf{0.997} & \textbf{0.999} & \textbf{0.050} & \textbf{0.148} & \textbf{2.072} & \textbf{0.076} \\
\end{tabular}
\vspace{1em}
\caption{\textbf{Depth estimation on the KITTI val set.}
Backbones pre-trained on ImageNet-22K are marked with $^\dagger$.
We report the performance of \ours with 3 diffusion steps.
The best and second-best results are bolded or underlined, respectively. ↓ means lower is better, and ↑ means higher is better. * denotes best results of our concurrent work~\cite{saxena2023depthgen}.}
\vspace{-1em}
\label{tab:exp:depth:kitti}
\end{table*}

\paragraph{Results.}
Table~\ref{tab:exp:depth:kitti} shows the depth estimation results on the KITTI dataset.
We compare the proposed \ours models with state-of-the-art depth estimators. 
Specifically, we choose DepthFormer \cite{li2022depthformer} and DepthGen \cite{saxena2023depthgen} as our main competitors, in which DepthFormer is a strong counterpart and achieved leading performance, while DepthGen is a concurrent work of ours and is also a diffusion-based depth estimator.
As we can observe, although the performance on this benchmark tends to be saturated,
our \ours models still outperform all the competitors with clear margins in most metrics, such as REL, SqRel, and RMSE.
For instance, equipped with Swin-L$^\dagger$, our method achieves a state-of-the-art RMSE log of 0.076 by 3 steps of denoising diffusion. 
Compared with the concurrent diffusion-based model~\cite{saxena2023depthgen}, we find that:
(1) \ours outperforms DepthGen with clear margins, particularly in regards to the RMSE$\downarrow$ metric (2.072 \emph{vs.} 2.985), which can be contributed by the equipped advanced pipeline design (e.g., Swin Transformer \emph{vs.} U-Net).
(2) \ours is more lightweight and efficient compared to DepthGen, as the denoising diffusion process occurs solely on the decoder head, whereas with DepthGen, the process occurs on the entire model.

\begin{table*}[t]
\centering
\hspace{-0.5em}
\subfloat[
    \textbf{Label encoding}. We find class embedding works best.
    \label{tab:label_encoding}
]{
    \begin{minipage}{0.18\linewidth}{
        \begin{center}
            \tablestyle{1pt}{1.0}
            \begin{tabular}{l|cc}
            \footnotesize{Type} & \footnotesize{mAcc} & \footnotesize{mIoU}  \\
            \hline
            \footnotesize{analog bits} & \footnotesize{57.6} & \footnotesize{46.2}  \\
            \footnotesize{onehot} & \footnotesize{56.8} & \footnotesize{46.2} \\
            \rowcolor{gray!10} 
            \footnotesize{\textbf{embedding}} &  \footnotesize{\textbf{58.4}} & \footnotesize{\textbf{47.0}}  \\
            \multicolumn{3}{c}{~}\\
            \multicolumn{3}{c}{~}\\
            \end{tabular}
        \end{center}}
    \end{minipage}
}
\hspace{1em}
\subfloat[
    \textbf{Scaling factor}. The best scaling factor is 0.01.
    \label{tab:scaling_factor}
]{
    \centering
    \begin{minipage}{0.155\linewidth}{
        \begin{center}
            \tablestyle{2pt}{1.0}
            \begin{tabular}{l|cc}
            \footnotesize{Scale} & \footnotesize{mAcc} & \footnotesize{mIoU} \\
            \hline
            \footnotesize{0.001} & \footnotesize{56.6} & \footnotesize{45.4} \\
            \rowcolor{gray!10} 
            \footnotesize{\textbf{0.01}} & \footnotesize{\textbf{58.4}} & \footnotesize{\textbf{47.0}} \\
            \footnotesize{0.02} & \footnotesize{57.5} & \footnotesize{46.8} \\
            \footnotesize{0.04} & \footnotesize{56.8} & \footnotesize{45.9} \\
            \footnotesize{0.1}  & \footnotesize{55.0} & \footnotesize{44.0} \\
            \end{tabular}
        \end{center}}
    \end{minipage}
}
\hspace{1em}
\subfloat[
    \textbf{Noise schedule}. Cosine works best.
    \label{tab:noise_schedule}
]{
    \begin{minipage}{0.13\linewidth}{
        \begin{center}
            \tablestyle{1pt}{1.0}
            \begin{tabular}{l|cc}
            \footnotesize{Type} & \footnotesize{mAcc} & \footnotesize{mIoU} \\
            \hline
            \rowcolor{gray!10} 
            \footnotesize{\textbf{cosine}} & \footnotesize{\textbf{58.4}} & \footnotesize{\textbf{47.0}} \\
            \footnotesize{linear} & \footnotesize{56.3} & \footnotesize{45.1} \\
            \multicolumn{3}{c}{~}\\
            \multicolumn{3}{c}{~}\\
            \multicolumn{3}{c}{~}\\
            \end{tabular}
        \end{center}}
    \end{minipage}
}
\hspace{1em}
\subfloat[
    \textbf{Decoder depth $L$}. Six blocks work best.
    \label{tab:decoder_depth}
]{
    \begin{minipage}{0.15\linewidth}{
        \begin{center}
            \tablestyle{1pt}{1.0}
            \begin{tabular}{l|ccc}
            \footnotesize{$L$} & \footnotesize{mAcc} & \footnotesize{mIoU}   & \footnotesize{\#Param} \\
            \hline
            \footnotesize{1} & \footnotesize{56.1} & \footnotesize{44.5} & \footnotesize{2.4M}  \\
            \footnotesize{2} & \footnotesize{56.5} & \footnotesize{45.0}& \footnotesize{3.6M}   \\
            \footnotesize{4} & \footnotesize{57.2} & \footnotesize{45.7} & \footnotesize{6.0M}  \\
            \rowcolor{gray!10} 
            \footnotesize{\textbf{6}} & \footnotesize{\textbf{58.4}} & \footnotesize{\textbf{47.0}} & \footnotesize{\textbf{8.4M}}  \\
            \footnotesize{12} & \footnotesize{55.7} & \footnotesize{46.0} & \footnotesize{15.6M}\\
            \end{tabular}
        \end{center}}
    \end{minipage}
}
\hspace{1em}
\subfloat[
    \textbf{Accuracy \emph{vs.}~Efficiency}. \colorbox{myyellow}{Yellow} denotes K-Net \cite{zhang2021k}.
    \label{tab:efficiency}
]{
    \begin{minipage}{0.18\linewidth}{
        \begin{center}
            \tablestyle{1.5pt}{1.0}
            \begin{tabular}{l|ccc}
            \footnotesize{Step} & \footnotesize{mIoU} &\footnotesize{FLOPs} & \footnotesize{FPS}  \\
            \hline
            \footnotesize{1}\cellcolor[HTML]{FFFADF} & \footnotesize{45.8}\cellcolor[HTML]{FFFADF} & \footnotesize{256G}\cellcolor[HTML]{FFFADF} & \footnotesize{18}\cellcolor[HTML]{FFFADF} \\
            \footnotesize{1} & \footnotesize{46.1} &\footnotesize{113G} & \footnotesize{19}  \\
            \footnotesize{2} & \footnotesize{46.8} & \footnotesize{182G} & \footnotesize{15}  \\
            \rowcolor{gray!10} 
            \footnotesize{\textbf{3}} & 
            \footnotesize{\textbf{47.0}} & \footnotesize{\textbf{252G}} & \footnotesize{\textbf{13}} \\
            
            \footnotesize{4} & \footnotesize{46.8} & \footnotesize{322G} &\footnotesize{11}  \\
            \end{tabular}
        \end{center}}
    \end{minipage}
}
\\
\caption{\textbf{\ours ablation experiments} with Swin-T~\cite{liu2021swin} on ADE20K semantic segmentation. We report the performance with 3 sampling steps in (a), (b), (c), and (d).
If not specified, the default settings are: 
the label encoding strategy is class embedding, the scaling factor is set to 0.01, the noise schedule is cosine, and the map decoder has a depth of 6. Default settings are marked in \colorbox{baselinecolor}{gray}.}

\label{tab:ablations}
\end{table*}

\subsection{Ablation Study}
We conduct ablation studies on the ADE20K semantic segmentation.
All models are trained using our \ours with Swin-T~\cite{liu2021swin} backbone for 160k iterations.
Other settings are the same as the settings in \cref{sec:semantic_settings}.
\label{sec:exp:abla}

\paragraph{Label Encoding.}
Since the labels of semantic segmentation are discrete, we need to encode them first. As shown in \cref{tab:label_encoding}, here we study the effect of three different strategies. For each of them, we search the optimal scaling factor. The results show that class embedding is a better strategy to encode semantic labels than one-hot and analog bits \cite{chen2022generalist}.

\paragraph{Signal Scale.}
As shown in \cref{tab:scaling_factor}, we search for the best scaling factor for the class embedding strategy. As can be seen, when we use a larger scaling factor than 0.01, the performance degraded significantly. This is because using a larger scaling factor, more easy cases are reserved with the same time step $t$.
In addition, we found the best scaling factor (\ie, 0.01) for class embedding is typically smaller than analog bits \cite{chen2022generalist} and one-hot (\ie, 0.1). 

\paragraph{Noise Schedule.} 
As shown in \cref{tab:noise_schedule}, we compare the effectiveness of the cosine schedule \cite{nichol2021improved} and linear schedule \cite{ho2020denoising} in \ours for semantic segmentation, and find that the model using the cosine schedule achieves notably better performance (47.0 \emph{vs.} 45.1).
This is attributed to the cosine schedule's mechanism of simulating the realistic scenario of gradually weakening signal influence, which prompts the model to learn stronger denoising capabilities, in contrast to the simple linear schedule.

\paragraph{Decoder Depth.}
We study the effect of decoder depth in \cref{tab:decoder_depth} and observe that the map decoder requires a suitable depth.
Initially, the model accuracy improves as the depth increases, but eventually decreases. 
Therefore, we finally adopted a map decoder with 6 blocks, which only has 8.4M parameters.
Overall, the map decoder is lightweight and efficient, compared with representative methods K-Net \cite{zhang2021k} (41.5M) and UperNet \cite{xiao2018unified} (31.5M).

\paragraph{Accuracy \emph{vs.} Efficiency.}

We show the dynamic trade-off of \ours between accuracy and efficiency  in \cref{tab:efficiency}. 
Compared with the representative discriminative method K-Net \cite{zhang2021k}, \ours yields a better mIoU when using only one sampling step, with fewer FLOPs and higher FPS. When adopting three sampling steps, the performance is further boosted to 47.0 mIoU, while maintaining comparable FLOPs and FPS.
These results show that \ours can iteratively infer multiple times with reasonable time cost.

\vspace{0.7em}
\section{Conclusion}
This paper introduced DDP, a simple, efficient, yet powerful framework for dense visual predictions based on conditional diffusion.
It extends the denoising diffusion process into modern perception pipelines, without requiring architectural customization or task-specific design.
We demonstrate DDP's effectiveness through state-of-the-art or competitive performance on three representative tasks and six diverse benchmarks.
Moreover, it additionally exhibits multiple inference and uncertainty awareness, which contrasts with previous single-step discriminative methods. 
These results indicate that DDP can serve as an important baseline for future research in dense prediction tasks.
One potential drawback of DDP is its non-negligible additional computational cost for multi-step inference. 
Besides, while DDP has demonstrated excellent improvement on several benchmark datasets for dense visual prediction tasks, further research is necessary to determine its efficacy in other domains.

\paragraph{Acknowledgement.}
We gratefully acknowledge the support of MindSpore, CANN (Compute Architecture for Neural Networks) and Ascend AI Processor used for this research.

{\small
\bibliographystyle{ieee_fullname}
\bibliography{src/bibs/egbib,src/bibs/diffusion}

\begin{thebibliography}{100}\itemsep=-1pt

\bibitem{amit2021segdiff}
Tomer Amit, Eliya Nachmani, Tal Shaharbany, and Lior Wolf.
\newblock Segdiff: Image segmentation with diffusion probabilistic models.
\newblock {\em arXiv preprint arXiv:2112.00390}, 2021.

\bibitem{anand2022protein}
Namrata Anand and Tudor Achim.
\newblock Protein structure and sequence generation with equivariant denoising
  diffusion probabilistic models.
\newblock {\em arXiv preprint arXiv:2205.15019}, 2022.

\bibitem{bhat2021adabins}
Shariq~Farooq Bhat, Ibraheem Alhashim, and Peter Wonka.
\newblock Adabins: Depth estimation using adaptive bins.
\newblock In {\em CVPR}, pages 4009--4018, 2021.

\bibitem{borse2023x}
Shubhankar Borse, Marvin Klingner, Varun~Ravi Kumar, Hong Cai, Abdulaziz
  Almuzairee, Senthil Yogamani, and Fatih Porikli.
\newblock X-align: Cross-modal cross-view alignment for bird's-eye-view
  segmentation.
\newblock In {\em WACV}, pages 3287--3297, 2023.

\bibitem{bousselham2021efficient}
Walid Bousselham, Guillaume Thibault, Lucas Pagano, Archana Machireddy, Joe
  Gray, Young~Hwan Chang, and Xubo Song.
\newblock Efficient self-ensemble framework for semantic segmentation.
\newblock {\em arXiv preprint arXiv:2111.13280}, 2021.

\bibitem{brock2019large}
Andrew Brock, Jeff Donahue, and Karen Simonyan.
\newblock Biggan: Large scale gan training for high fidelity natural image
  synthesis.
\newblock In {\em ICLR}, 2019.

\bibitem{caesar2020nuscenes}
Holger Caesar, Varun Bankiti, Alex~H Lang, Sourabh Vora, Venice~Erin Liong,
  Qiang Xu, Anush Krishnan, Yu Pan, Giancarlo Baldan, and Oscar Beijbom.
\newblock nuscenes: A multimodal dataset for autonomous driving.
\newblock In {\em CVPR}, pages 11621--11631, 2020.

\bibitem{mmdetection}
Kai Chen, Jiaqi Wang, Jiangmiao Pang, Yuhang Cao, Yu Xiong, Xiaoxiao Li,
  Shuyang Sun, Wansen Feng, Ziwei Liu, Jiarui Xu, Zheng Zhang, Dazhi Cheng,
  Chenchen Zhu, Tianheng Cheng, Qijie Zhao, Buyu Li, Xin Lu, Rui Zhu, Yue Wu,
  Jifeng Dai, Jingdong Wang, Jianping Shi, Wanli Ouyang, Chen~Change Loy, and
  Dahua Lin.
\newblock {MMDetection}: Open mmlab detection toolbox and benchmark.
\newblock {\em arXiv preprint arXiv:1906.07155}, 2019.

\bibitem{chen2017deeplab}
Liang-Chieh Chen, George Papandreou, Iasonas Kokkinos, Kevin Murphy, and Alan~L
  Yuille.
\newblock Deeplab: Semantic image segmentation with deep convolutional nets,
  atrous convolution, and fully connected crfs.
\newblock {\em TPAMI}, 2017.

\bibitem{chen2018encoder}
Liang-Chieh Chen, Yukun Zhu, George Papandreou, Florian Schroff, and Hartwig
  Adam.
\newblock Encoder-decoder with atrous separable convolution for semantic image
  segmentation.
\newblock In {\em ECCV}, pages 801--818, 2018.

\bibitem{chen2022sampling}
Sitan Chen, Sinho Chewi, Jerry Li, Yuanzhi Li, Adil Salim, and Anru~R Zhang.
\newblock Sampling is as easy as learning the score: theory for diffusion
  models with minimal data assumptions.
\newblock {\em arXiv preprint arXiv:2209.11215}, 2022.

\bibitem{chen2022diffusiondet}
Shoufa Chen, Peize Sun, Yibing Song, and Ping Luo.
\newblock Diffusiondet: Diffusion model for object detection.
\newblock {\em arXiv preprint arXiv:2211.09788}, 2022.

\bibitem{chen2023importance}
Ting Chen.
\newblock On the importance of noise scheduling for diffusion models.
\newblock {\em arXiv preprint arXiv:2301.10972}, 2023.

\bibitem{chen2022generalist}
Ting Chen, Lala Li, Saurabh Saxena, Geoffrey Hinton, and David~J Fleet.
\newblock A generalist framework for panoptic segmentation of images and
  videos.
\newblock {\em arXiv preprint arXiv:2210.06366}, 2022.

\bibitem{chen2022analog}
Ting Chen, Ruixiang Zhang, and Geoffrey Hinton.
\newblock Analog bits: Generating discrete data using diffusion models with
  self-conditioning.
\newblock {\em arXiv preprint arXiv:2208.04202}, 2022.

\bibitem{chen2023analog}
Ting Chen, Ruixiang Zhang, and Geoffrey Hinton.
\newblock Analog bits: Generating discrete data using diffusion models with
  self-conditioning.
\newblock In {\em ICLR}, 2023.

\bibitem{chen2022vitadapter}
Zhe Chen, Yuchen Duan, Wenhai Wang, Junjun He, Tong Lu, Jifeng Dai, and Yu
  Qiao.
\newblock Vision transformer adapter for dense predictions.
\newblock In {\em ICLR}, 2023.

\bibitem{cheng2022masked}
Bowen Cheng, Ishan Misra, Alexander~G Schwing, Alexander Kirillov, and Rohit
  Girdhar.
\newblock Masked-attention mask transformer for universal image segmentation.
\newblock In {\em CVPR}, pages 1290--1299, 2022.

\bibitem{cheng2021per}
Bowen Cheng, Alex Schwing, and Alexander Kirillov.
\newblock Per-pixel classification is not all you need for semantic
  segmentation.
\newblock {\em NeurIPS}, 34:17864--17875, 2021.

\bibitem{mmseg2020}
MMSegmentation Contributors.
\newblock {MMSegmentation}: Openmmlab semantic segmentation toolbox and
  benchmark.
\newblock \url{https://github.com/open-mmlab/mmsegmentation}, 2020.

\bibitem{Cordts_2016_CVPR}
Marius Cordts, Mohamed Omran, Sebastian Ramos, Timo Rehfeld, Markus Enzweiler,
  Rodrigo Benenson, Uwe Franke, Stefan Roth, and Bernt Schiele.
\newblock The cityscapes dataset for semantic urban scene understanding.
\newblock In {\em CVPR}, 2016.

\bibitem{corso2022diffdock}
Gabriele Corso, Hannes St{\"a}rk, Bowen Jing, Regina Barzilay, and Tommi
  Jaakkola.
\newblock Diffdock: Diffusion steps, twists, and turns for molecular docking.
\newblock {\em arXiv preprint arXiv:2210.01776}, 2022.

\bibitem{cui2022region}
Jiequan Cui, Yuhui Yuan, Zhisheng Zhong, Zhuotao Tian, Han Hu, Stephen Lin, and
  Jiaya Jia.
\newblock Region rebalance for long-tailed semantic segmentation.
\newblock {\em arXiv preprint arXiv:2204.01969}, 2022.

\bibitem{daras2023consistent}
Giannis Daras, Yuval Dagan, Alexandros~G Dimakis, and Constantinos Daskalakis.
\newblock Consistent diffusion models: Mitigating sampling drift by learning to
  be consistent.
\newblock {\em arXiv preprint arXiv:2302.09057}, 2023.

\bibitem{daras2022multiresolution}
Giannis Daras and Alexandros~G Dimakis.
\newblock Multiresolution textual inversion.
\newblock {\em arXiv preprint arXiv:2211.17115}, 2022.

\bibitem{deng2009imagenet}
Jia Deng, Wei Dong, Richard Socher, Li-Jia Li, Kai Li, and Li Fei-Fei.
\newblock Imagenet: A large-scale hierarchical image database.
\newblock In {\em CVPR}, pages 248--255, 2009.

\bibitem{dhariwal2021diffusion}
Prafulla Dhariwal and Alexander Nichol.
\newblock Diffusion models beat gans on image synthesis.
\newblock {\em NeurIPS}, 34:8780--8794, 2021.

\bibitem{eigen2014depth}
David Eigen, Christian Puhrsch, and Rob Fergus.
\newblock Depth map prediction from a single image using a multi-scale deep
  network.
\newblock {\em NeurIPS}, 2014.

\bibitem{fischer2015flownet}
Philipp Fischer, Alexey Dosovitskiy, Eddy Ilg, Philip H{\"a}usser, Caner
  Haz{\i}rba{\c{s}}, Vladimir Golkov, Patrick Van~der Smagt, Daniel Cremers,
  and Thomas Brox.
\newblock Flownet: Learning optical flow with convolutional networks.
\newblock {\em arXiv preprint arXiv:1504.06852}, 2015.

\bibitem{fu2018deep}
Huan Fu, Mingming Gong, Chaohui Wang, Kayhan Batmanghelich, and Dacheng Tao.
\newblock Deep ordinal regression network for monocular depth estimation.
\newblock In {\em CVPR}, pages 2002--2011, 2018.

\bibitem{geiger2013vision}
Andreas Geiger, Philip Lenz, Christoph Stiller, and Raquel Urtasun.
\newblock Vision meets robotics: The kitti dataset.
\newblock {\em IJRR}, 32(11):1231--1237, 2013.

\bibitem{gong2021vision}
Chengyue Gong, Dilin Wang, Meng Li, Vikas Chandra, and Qiang Liu.
\newblock Vision transformers with patch diversification.
\newblock {\em arXiv preprint arXiv:2104.12753}, 2021.

\bibitem{harakeh2020bayesod}
Ali Harakeh, Michael Smart, and Steven~L Waslander.
\newblock Bayesod: A bayesian approach for uncertainty estimation in deep
  object detectors.
\newblock In {\em ICRA}, pages 87--93. IEEE, 2020.

\bibitem{hendrik2017universal}
Jan Hendrik~Metzen, Mummadi Chaithanya~Kumar, Thomas Brox, and Volker Fischer.
\newblock Universal adversarial perturbations against semantic image
  segmentation.
\newblock In {\em ICCV}, pages 2755--2764, 2017.

\bibitem{ho2020denoising}
Jonathan Ho, Ajay Jain, and Pieter Abbeel.
\newblock Denoising diffusion probabilistic models.
\newblock {\em NeurIPS}, 33:6840--6851, 2020.

\bibitem{ho2022video}
Jonathan Ho, Tim Salimans, Alexey Gritsenko, William Chan, Mohammad Norouzi,
  and David~J Fleet.
\newblock Video diffusion models.
\newblock {\em arXiv:2204.03458}, 2022.

\bibitem{hong2022cogvideo}
Wenyi Hong, Ming Ding, Wendi Zheng, Xinghan Liu, and Jie Tang.
\newblock Cogvideo: Large-scale pretraining for text-to-video generation via
  transformers.
\newblock {\em arXiv preprint arXiv:2205.15868}, 2022.

\bibitem{huynh2020guiding}
Lam Huynh, Phong Nguyen-Ha, Jiri Matas, Esa Rahtu, and Janne Heikkil{\"a}.
\newblock Guiding monocular depth estimation using depth-attention volume.
\newblock In {\em ECCV}, pages 581--597, 2020.

\bibitem{isola2017image}
Phillip Isola, Jun-Yan Zhu, Tinghui Zhou, and Alexei~A Efros.
\newblock Image-to-image translation with conditional adversarial networks.
\newblock In {\em CVPR}, pages 1125--1134, 2017.

\bibitem{jain2022oneformer}
Jitesh Jain, Jiachen Li, MangTik Chiu, Ali Hassani, Nikita Orlov, and Humphrey
  Shi.
\newblock Oneformer: One transformer to rule universal image segmentation.
\newblock {\em arXiv preprint arXiv:2211.06220}, 2022.

\bibitem{ji2021monoindoor}
Pan Ji, Runze Li, Bir Bhanu, and Yi Xu.
\newblock Monoindoor: Towards good practice of self-supervised monocular depth
  estimation for indoor environments.
\newblock In {\em ICCV}, pages 12787--12796, 2021.

\bibitem{karras2018progressive}
Tero Karras, Timo Aila, Samuli Laine, and Jaakko Lehtinen.
\newblock Progressive growing of gans for improved quality, stability, and
  variation.
\newblock In {\em ICLR}, 2018.

\bibitem{kolesnikov2020big}
Alexander Kolesnikov, Lucas Beyer, Xiaohua Zhai, Joan Puigcerver, Jessica Yung,
  Sylvain Gelly, and Neil Houlsby.
\newblock Big transfer (bit): General visual representation learning.
\newblock In {\em ECCV}, pages 491--507, 2020.

\bibitem{lee2019big}
Jin~Han Lee, Myung-Kyu Han, Dong~Wook Ko, and Il~Hong Suh.
\newblock From big to small: Multi-scale local planar guidance for monocular
  depth estimation.
\newblock {\em arXiv preprint arXiv:1907.10326}, 2019.

\bibitem{li2021structdepth}
Boying Li, Yuan Huang, Zeyu Liu, Danping Zou, and Wenxian Yu.
\newblock Structdepth: Leveraging the structural regularities for
  self-supervised indoor depth estimation.
\newblock In {\em ICCV}, pages 12663--12673, 2021.

\bibitem{li2021semantic}
Daiqing Li, Junlin Yang, Karsten Kreis, Antonio Torralba, and Sanja Fidler.
\newblock Semantic segmentation with generative models: Semi-supervised
  learning and strong out-of-domain generalization.
\newblock In {\em CVPR}, pages 8300--8311, 2021.

\bibitem{li2022uniformer}
Kunchang Li, Yali Wang, Peng Gao, Guanglu Song, Yu Liu, Hongsheng Li, and Yu
  Qiao.
\newblock Uniformer: Unified transformer for efficient spatiotemporal
  representation learning.
\newblock {\em arXiv preprint arXiv:2201.04676}, 2022.

\bibitem{li2022depthformer}
Zhenyu Li, Zehui Chen, Xianming Liu, and Junjun Jiang.
\newblock Depthformer: Exploiting long-range correlation and local information
  for accurate monocular depth estimation.
\newblock {\em arXiv preprint arXiv:2203.14211}, 2022.

\bibitem{li2022binsformer}
Zhenyu Li, Xuyang Wang, Xianming Liu, and Junjun Jiang.
\newblock Binsformer: Revisiting adaptive bins for monocular depth estimation.
\newblock {\em arXiv preprint arXiv:2204.00987}, 2022.

\bibitem{lin2022structtoken}
Fangjian Lin, Zhanhao Liang, Junjun He, Miao Zheng, Shengwei Tian, and Kai
  Chen.
\newblock Structtoken: Rethinking semantic segmentation with structural prior.
\newblock {\em arXiv preprint arXiv:2203.12612}, 2022.

\bibitem{lin2017feature}
Tsung-Yi Lin, Piotr Doll{\'a}r, Ross Girshick, Kaiming He, Bharath Hariharan,
  and Serge Belongie.
\newblock Feature pyramid networks for object detection.
\newblock In {\em CVPR}, pages 2117--2125, 2017.

\bibitem{liu2021swin}
Ze Liu, Yutong Lin, Yue Cao, Han Hu, Yixuan Wei, Zheng Zhang, Stephen Lin, and
  Baining Guo.
\newblock Swin transformer: Hierarchical vision transformer using shifted
  windows.
\newblock In {\em ICCV}, pages 10012--10022, 2021.

\bibitem{liu2022convnet}
Zhuang Liu, Hanzi Mao, Chao-Yuan Wu, Christoph Feichtenhofer, Trevor Darrell,
  and Saining Xie.
\newblock A convnet for the 2020s.
\newblock {\em arXiv preprint arXiv:2201.03545}, 2022.

\bibitem{liu2022bevfusion}
Zhijian Liu, Haotian Tang, Alexander Amini, Xinyu Yang, Huizi Mao, Daniela Rus,
  and Song Han.
\newblock Bevfusion: Multi-task multi-sensor fusion with unified bird's-eye
  view representation.
\newblock {\em arXiv preprint arXiv:2205.13542}, 2022.

\bibitem{loquercio2020general}
Antonio Loquercio, Mattia Segu, and Davide Scaramuzza.
\newblock A general framework for uncertainty estimation in deep learning.
\newblock {\em IEEE Robotics and Automation Letters}, 5(2):3153--3160, 2020.

\bibitem{loshchilov2017decoupled}
Ilya Loshchilov and Frank Hutter.
\newblock Decoupled weight decay regularization.
\newblock {\em arXiv preprint arXiv:1711.05101}, 2017.

\bibitem{nichol2021glide}
Alex Nichol, Prafulla Dhariwal, Aditya Ramesh, Pranav Shyam, Pamela Mishkin,
  Bob McGrew, Ilya Sutskever, and Mark Chen.
\newblock Glide: Towards photorealistic image generation and editing with
  text-guided diffusion models.
\newblock {\em arXiv preprint arXiv:2112.10741}, 2021.

\bibitem{nichol2021improved}
Alexander~Quinn Nichol and Prafulla Dhariwal.
\newblock Improved denoising diffusion probabilistic models.
\newblock In {\em ICML}, pages 8162--8171, 2021.

\bibitem{philion2020lift}
Jonah Philion and Sanja Fidler.
\newblock Lift, splat, shoot: Encoding images from arbitrary camera rigs by
  implicitly unprojecting to 3d.
\newblock In {\em ECCV}, pages 194--210, 2020.

\bibitem{ramesh2022hierarchical}
Aditya Ramesh, Prafulla Dhariwal, Alex Nichol, Casey Chu, and Mark Chen.
\newblock Hierarchical text-conditional image generation with clip latents.
\newblock {\em arXiv preprint arXiv:2204.06125}, 2022.

\bibitem{ranftl2021vision}
Ren{\'e} Ranftl, Alexey Bochkovskiy, and Vladlen Koltun.
\newblock Vision transformers for dense prediction.
\newblock In {\em ICCV}, pages 12179--12188, 2021.

\bibitem{roddick2018orthographic}
Thomas Roddick, Alex Kendall, and Roberto Cipolla.
\newblock Orthographic feature transform for monocular 3d object detection.
\newblock {\em arXiv preprint arXiv:1811.08188}, 2018.

\bibitem{rombach2022high}
Robin Rombach, Andreas Blattmann, Dominik Lorenz, Patrick Esser, and Bj{\"o}rn
  Ommer.
\newblock High-resolution image synthesis with latent diffusion models.
\newblock In {\em CVPR}, pages 10684--10695, 2022.

\bibitem{ronneberger2015u}
Olaf Ronneberger, Philipp Fischer, and Thomas Brox.
\newblock U-net: Convolutional networks for biomedical image segmentation.
\newblock In {\em MICCAI}, pages 234--241, 2015.

\bibitem{saharia2022photorealistic}
Chitwan Saharia, William Chan, Saurabh Saxena, Lala Li, Jay Whang, Emily
  Denton, Seyed Kamyar~Seyed Ghasemipour, Burcu~Karagol Ayan, S~Sara Mahdavi,
  Rapha~Gontijo Lopes, et~al.
\newblock Photorealistic text-to-image diffusion models with deep language
  understanding.
\newblock {\em arXiv preprint arXiv:2205.11487}, 2022.

\bibitem{saharia2022image}
Chitwan Saharia, Jonathan Ho, William Chan, Tim Salimans, David~J Fleet, and
  Mohammad Norouzi.
\newblock Image super-resolution via iterative refinement.
\newblock {\em TPAMI}, 2022.

\bibitem{salimans2016improved}
Tim Salimans, Ian Goodfellow, Wojciech Zaremba, Vicki Cheung, Alec Radford, and
  Xi Chen.
\newblock Improved techniques for training gans.
\newblock {\em NeurIPS}, 29:2234--2242, 2016.

\bibitem{saxena2023depthgen}
Saurabh Saxena, Abhishek Kar, Mohammad Norouzi, and David~J. Fleet.
\newblock Monocular depth estimation using diffusion models.
\newblock {\em arXiv preprint arXiv:2302.14816}, 2023.

\bibitem{schneuing2022structure}
Arne Schneuing, Yuanqi Du, Charles Harris, Arian Jamasb, Ilia Igashov, Weitao
  Du, Tom Blundell, Pietro Li{\'o}, Carla Gomes, Max Welling, et~al.
\newblock Structure-based drug design with equivariant diffusion models.
\newblock {\em arXiv preprint arXiv:2210.13695}, 2022.

\bibitem{silberman2012indoor}
Nathan Silberman, Derek Hoiem, Pushmeet Kohli, and Rob Fergus.
\newblock Indoor segmentation and support inference from rgbd images.
\newblock In {\em ECCV}, pages 746--760, 2012.

\bibitem{sohl2015deep}
Jascha Sohl-Dickstein, Eric Weiss, Niru Maheswaranathan, and Surya Ganguli.
\newblock Deep unsupervised learning using nonequilibrium thermodynamics.
\newblock In {\em ICML}, pages 2256--2265, 2015.

\bibitem{song2020denoising}
Jiaming Song, Chenlin Meng, and Stefano Ermon.
\newblock Denoising diffusion implicit models.
\newblock {\em arXiv preprint arXiv:2010.02502}, 2020.

\bibitem{songdenoising}
Jiaming Song, Chenlin Meng, and Stefano Ermon.
\newblock Denoising diffusion implicit models.
\newblock In {\em ICLR}, 2021.

\bibitem{song2015sun}
Shuran Song, Jianxiong Xiao, Li Guo, and Xiaogang Yang.
\newblock Sun rgb-d: A rgb-d scene understanding benchmark suite.
\newblock {\em CVPR}, 2015.

\bibitem{song2020improved}
Yang Song and Stefano Ermon.
\newblock Improved techniques for training score-based generative models.
\newblock {\em NeurIPS}, 33:12438--12448, 2020.

\bibitem{strudel2021segmenter}
Robin Strudel, Ricardo Garcia, Ivan Laptev, and Cordelia Schmid.
\newblock Segmenter: Transformer for semantic segmentation.
\newblock In {\em ICCV}, pages 7262--7272, 2021.

\bibitem{trippe2022diffusion}
Brian~L Trippe, Jason Yim, Doug Tischer, Tamara Broderick, David Baker, Regina
  Barzilay, and Tommi Jaakkola.
\newblock Diffusion probabilistic modeling of protein backbones in 3d for the
  motif-scaffolding problem.
\newblock {\em arXiv preprint arXiv:2206.04119}, 2022.

\bibitem{vora2020pointpainting}
Sourabh Vora, Alex~H Lang, Bassam Helou, and Oscar Beijbom.
\newblock Pointpainting: Sequential fusion for 3d object detection.
\newblock In {\em CVPR}, pages 4604--4612, 2020.

\bibitem{wang2022internimage}
Wenhai Wang, Jifeng Dai, Zhe Chen, Zhenhang Huang, Zhiqi Li, Xizhou Zhu,
  Xiaowei Hu, Tong Lu, Lewei Lu, Hongsheng Li, et~al.
\newblock Internimage: Exploring large-scale vision foundation models with
  deformable convolutions.
\newblock {\em arXiv preprint arXiv:2211.05778}, 2022.

\bibitem{wang2021pyramid}
Wenhai Wang, Enze Xie, Xiang Li, Deng-Ping Fan, Kaitao Song, Ding Liang, Tong
  Lu, Ping Luo, and Ling Shao.
\newblock Pyramid vision transformer: A versatile backbone for dense prediction
  without convolutions.
\newblock In {\em ICCV}, pages 568--578, 2021.

\bibitem{wolleb2022diffusion}
Julia Wolleb, Robin Sandk{\"u}hler, Florentin Bieder, Philippe Valmaggia, and
  Philippe~C Cattin.
\newblock Diffusion models for implicit image segmentation ensembles.
\newblock In {\em MIDL}, pages 1336--1348, 2022.

\bibitem{wu2022medsegdiff}
Junde Wu, Huihui Fang, Yu Zhang, Yehui Yang, and Yanwu Xu.
\newblock Medsegdiff: Medical image segmentation with diffusion probabilistic
  model.
\newblock {\em arXiv preprint arXiv:2211.00611}, 2022.

\bibitem{xiao2018unified}
Tete Xiao, Yingcheng Liu, Bolei Zhou, Yuning Jiang, and Jian Sun.
\newblock Unified perceptual parsing for scene understanding.
\newblock In {\em ECCV}, pages 418--434, 2018.

\bibitem{xie2017adversarial}
Cihang Xie, Jianyu Wang, Zhishuai Zhang, Yuyin Zhou, Lingxi Xie, and Alan
  Yuille.
\newblock Adversarial examples for semantic segmentation and object detection.
\newblock In {\em ICCV}, pages 1369--1378, 2017.

\bibitem{xie2021segformer}
Enze Xie, Wenhai Wang, Zhiding Yu, Anima Anandkumar, Jose~M Alvarez, and Ping
  Luo.
\newblock Segformer: Simple and efficient design for semantic segmentation with
  transformers.
\newblock {\em NeurIPS}, 34, 2021.

\bibitem{xie2022m2bev}
Enze Xie, Zhiding Yu, Daquan Zhou, Jonah Philion, Anima Anandkumar, Sanja
  Fidler, Ping Luo, and Jose~M Alvarez.
\newblock M\^{2} bev: Multi-camera joint 3d detection and segmentation with
  unified birds-eye view representation.
\newblock {\em arXiv preprint arXiv:2204.05088}, 2022.

\bibitem{yang2021transformer}
Guanglei Yang, Hao Tang, Mingli Ding, Nicu Sebe, and Elisa Ricci.
\newblock Transformer-based attention networks for continuous pixel-wise
  prediction.
\newblock In {\em ICCV}, pages 16269--16279, 2021.

\bibitem{yeh2017semantic}
Raymond~A Yeh, Chen Chen, Teck Yian~Lim, Alexander~G Schwing, Mark
  Hasegawa-Johnson, and Minh~N Do.
\newblock Semantic image inpainting with deep generative models.
\newblock In {\em CVPR}, pages 5485--5493, 2017.

\bibitem{yin2021multimodal}
Tianwei Yin, Xingyi Zhou, and Philipp Kr{\"a}henb{\"u}hl.
\newblock Multimodal virtual point 3d detection.
\newblock {\em NeurIPS}, 34:16494--16507, 2021.

\bibitem{yin2019enforcing}
Wei Yin, Yifan Liu, Chunhua Shen, and Youliang Yan.
\newblock Enforcing geometric constraints of virtual normal for depth
  prediction.
\newblock In {\em ICCV}, pages 5684--5693, 2019.

\bibitem{yuan2020object}
Yuhui Yuan, Xilin Chen, and Jingdong Wang.
\newblock Object-contextual representations for semantic segmentation.
\newblock In {\em ECCV}, pages 173--190, 2020.

\bibitem{yuan2021hrformer}
Yuhui Yuan, Rao Fu, Lang Huang, Weihong Lin, Chao Zhang, Xilin Chen, and
  Jingdong Wang.
\newblock Hrformer: High-resolution vision transformer for dense prediction.
\newblock {\em NeurIPS}, 34, 2021.

\bibitem{zhang2022dino}
Hao Zhang, Feng Li, Shilong Liu, Lei Zhang, Hang Su, Jun Zhu, Lionel~M Ni, and
  Heung-Yeung Shum.
\newblock Dino: Detr with improved denoising anchor boxes for end-to-end object
  detection.
\newblock {\em arXiv preprint arXiv:2203.03605}, 2022.

\bibitem{zhang2023adding}
Lvmin Zhang and Maneesh Agrawala.
\newblock Adding conditional control to text-to-image diffusion models.
\newblock {\em arXiv preprint arXiv:2302.05543}, 2023.

\bibitem{zhang2021k}
Wenwei Zhang, Jiangmiao Pang, Kai Chen, and Chen~Change Loy.
\newblock K-net: Towards unified image segmentation.
\newblock {\em NeurIPS}, pages 10326--10338, 2021.

\bibitem{zhao2017pyramid}
Hengshuang Zhao, Jianping Shi, Xiaojuan Qi, Xiaogang Wang, and Jiaya Jia.
\newblock Pyramid scene parsing network.
\newblock In {\em CVPR}, pages 2881--2890, 2017.

\bibitem{zheng2021rethinking}
Sixiao Zheng, Jiachen Lu, Hengshuang Zhao, Xiatian Zhu, Zekun Luo, Yabiao Wang,
  Yanwei Fu, Jianfeng Feng, Tao Xiang, Philip~HS Torr, et~al.
\newblock Rethinking semantic segmentation from a sequence-to-sequence
  perspective with transformers.
\newblock In {\em CVPR}, pages 6881--6890, 2021.

\bibitem{zhou2022cross}
Brady Zhou and Philipp Kr{\"a}henb{\"u}hl.
\newblock Cross-view transformers for real-time map-view semantic segmentation.
\newblock In {\em CVPR}, pages 13760--13769, 2022.

\bibitem{zhou2017scene}
Bolei Zhou, Hang Zhao, Xavier Puig, Sanja Fidler, Adela Barriuso, and Antonio
  Torralba.
\newblock Scene parsing through ade20k dataset.
\newblock In {\em CVPR}, pages 633--641, 2017.

\bibitem{zhu2020deformable}
Xizhou Zhu, Weijie Su, Lewei Lu, Bin Li, Xiaogang Wang, and Jifeng Dai.
\newblock Deformable detr: Deformable transformers for end-to-end object
  detection.
\newblock 2020.

\end{thebibliography}
}

\clearpage

\appendix

\section{Diffusion Model}
\label{supp:diffusion}

\subsection{Algorithm details}

\begin{figure}[t!]
    \centering
    \includegraphics[width=\linewidth]{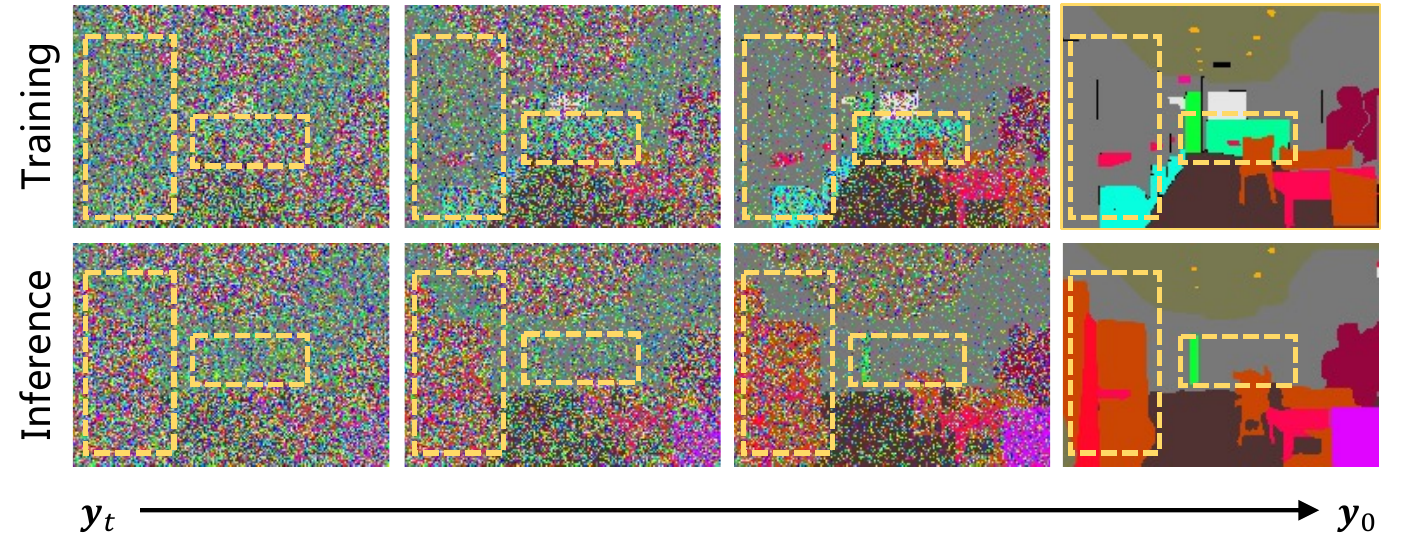}
    \caption{\textbf{Sampling drift}. Denoising targets differ from the training process and inference process. }
    \label{fig:dis:shift}
\end{figure}

As a supplement to \cref{algo:ddp:training} and \cref{algo:ddp:sampling}  described in the main paper, we provide the implementation details in \cref{algo:ddp:ddim} for better clarity.
Additionally, we introduce the implementation of the ``\emph{self-aligned denoising}" procedure in \cref{algo:ddp:self}, used in the last 5K iteration training to address the sampling drift problem (see \cref{sec:inference}).
We provide an example  in \cref{fig:dis:shift} to illustrate the  gap between the training and inference denoising targets.

\begin{algorithm}[t!]
\caption{DDIM Update}
\label{algo:ddp:ddim}
\definecolor{codeblue}{HTML}{2E8B57} 
\definecolor{codekw}{HTML}{DC143C} 
\lstset{
  backgroundcolor=\color{white},
  columns=fullflexible,
  breaklines=true,
  captionpos=b,
  commentstyle=\fontsize{7.5pt}{7.5pt}\color{codeblue},
  keywordstyle=\fontsize{7.5pt}{7.5pt}\color{codekw},
  escapechar={|}, 
}
\lstset{language=Python}
\begin{lstlisting}[mathescape,xleftmargin=-1em]
def alpha_cumprod(t, ns=0.0002, ds=0.00025):
 """cosine noise schedule"""
 n = torch.cos((t + ns) / (1 + ds)
     * math.pi / 2) ** -2
 return -torch.log(n - 1, eps=1e-5)
 
def ddim(map_t, map_pred, t_now, t_next):
  """
  estimate x at t_next with DDIM update rule.
  """
  $\alpha_\text{now}$ = alpha_cumprod(t_now)
  $\alpha_\text{next}$ = alpha_cumprod(t_next)
  map_enc = encoding(map_pred)
  map_enc = (sigmoid(map_enc) * 2 - 1) * scale
  eps = $\frac{1}{\sqrt{1-\alpha_{\text{now}}}}$ * (map_t - $\sqrt{\alpha_{\text{now}}}$ * map_enc)
  map_next = $\sqrt{\alpha_{\text{next}}}$ * x_pred + $\sqrt{1-\alpha_{\text{now}}}$ * eps
  return map_next
\end{lstlisting}
\vspace{0.475em}
\end{algorithm}

\subsection{More Discussions}

As illustrated in \cref{fig:multiple_inference}, diffusion models for perceptual tasks tend to reach a saturation point within the first few steps, usually between 3-5 steps, making additional diffusion less advantageous. This is in contrast to the requirements of generative models for image generation, where multiple iterations over many steps (from 10 to 50) are often necessary.
Intuitively, in generative tasks such as image generation, the goal is to produce complete and high-quality results by progressively incorporating more information at each time step, thus gradually accumulating and improving the overall result.
Therefore, it may take more time steps to reach convergence in order to fully accumulate the necessary information.
In perceptual tasks, such as semantic segmentation and object detection, the process from image to label is a gradual reduction of information, and critical information sufficient to make a decision needs to be obtained in only a few steps. Therefore, further diffusion has a limited role in improving the accuracy of predictions, leading to an early peak within three to five steps.
In short, the diffusion process in a perception task can make decisions by accumulating the most important information. Therefore, DDP can achieve high accuracy in perception tasks with minimal computational cost.

\begin{algorithm}[t!]
\caption{ \ours Self-aligned Denoising}
\label{algo:ddp:self}
\definecolor{codeblue}{HTML}{2E8B57} 
\definecolor{codekw}{HTML}{DC143C}
\lstset{
  backgroundcolor=\color{white},
  columns=fullflexible,
  breaklines=true,
  captionpos=b,
  commentstyle=\fontsize{7.2pt}{7.2pt}\color{codeblue},
  keywordstyle=\fontsize{7.2pt}{7.2pt}\color{codekw},
  escapechar={|}, 
}
\lstset{language=Python}
\begin{lstlisting}[mathescape,xleftmargin=-1em]
def train(images, maps):
  """
  images: [b, 3, h, w], maps: [b, 1, h, w]
  """
  img_enc = image_encoder(images)
  map_t = normal(mean=0, std=1)
  map_pred = map_decoder(map_t, img_enc, t=1)
  # encode map_pred
  map_enc = encoding(map_pred.detach())
  map_enc = (sigmoid(map_enc) * 2 - 1) * scale
  # corrupt the map_enc
  t, eps = uniform(0, 1), normal(mean=0, std=1)
  map_crpt = sqrt(alpha_cumprod(t)) * map_enc +
              sqrt(1 - alpha_cumprod(t)) * eps
  # predict
  map_pred = map_decoder(map_crpt, img_enc, t)
  loss = objective_func(map_pred, maps)
  return loss
\end{lstlisting}
\end{algorithm}

\section{Implementation Details}
\label{supp:implement}
\subsection{Semantic Segmentation}
\paragraph{ADE20K.} We conduct the experiments of ADE20K \cite{zhou2017scene} semantic segmentation based on MMSegmentation~\cite{mmseg2020}.
In the training phase, the backbone is initialized with the ImageNet \cite{deng2009imagenet} pre-trained weights.
We optimize our \ours models using AdamW~\cite{loshchilov2017decoupled} optimizer with an initial learning rate of 6$\times10^{-5}$, and a weight decay of 0.01.
The learning rate is decayed following the polynomial decay schedule with a power of 1.0.
Besides, we randomly resize and crop the image to 512$\times$512 for training, and rescale to have a shorter side of 512 pixels during testing.
All models are trained for 160k iterations with a batch size of 16 and compared fairly with previous discriminative-based and non-diffusion methods.

\paragraph{Cityscapes.}
The Cityscape dataset includes 5000 high-resolution images, which contain  2,975 training images, 500 validation images, and 1525 testing samples.
The images are captured from 50 different cities in Germany, covering various environments such as highways, city centers, and suburbs.
Similar to ADE20K, during training, we load the ImageNet pre-trained weights and employ the AdamW optimizer.
Following common practice, we randomly resize and crop the image to 512$\times$1024 for training, and take the original images of 1024$\times$2048 for testing.
We 
Other hyper-parameters are kept the same as our ADE20K experiments.

\subsection{BEV Map Segmentation}
\paragraph{nuScenes.}
We conduct our experiments of BEV map segmentation on nuScenes~\cite{caesar2020nuscenes}, a large-scale multi-modal dataset for 3D detection and map segmentation.
The dataset is split into 700/150/150 scenes for training/validation/testing.
It contains data from multiple sensors, including six cameras, one LIDAR, and five radars.
For camera inputs, each frame consists of six views of the surrounding  environment at the same timestamps.
We resize the input views to 256$\times$704 and voxelize the point cloud to 0.1m.
Our evaluation metrics align with \cite{liu2022bevfusion} and report the IoU of 6 background classes, including drivable space, pedestrian crossing, walk-way, stop line, car-parking area, and lane divider, and use the mean IoU as the primary evaluation metric.
We adopt the image and LiDAR data augmentation strategies from \cite{mmdetection} for training.
AdamW is utilized with a weight decay of 0.01 and a learning rate of 5e-5.
We take overall 20 training epochs on 8 A100 GPUs with a batch size of 32.
Other training settings are kept the same as \cite{liu2022bevfusion} for fair comparisons.
\subsection{Depth Estimation}
\paragraph{KITTI.}
The KITTI depth estimation dataset is a widely used benchmark dataset for monocular depth estimation with a depth range from 0-80m.
The stereo images of the dataset have a resolution of 1242$\times$375, while the corresponding GT depth map has a low density of 3.75\% to 5.0\%. 
Following the standard Eigen training/testing split~\cite{eigen2014depth}, we use around 26K left view images for training and 697 frames for testing.
We incorporate the \ours model into the codebase developed by \cite{li2022depthformer} for KITTI depth estimation experiments.
We excluded the discrete label encoding module as the task requires continuous value regression
All experimental settings are the same as \cite{li2022depthformer} for a fair comparison.

\begin{table}[t!]
\centering
\footnotesize
\setlength{\tabcolsep}{1.25mm}
\begin{tabular}{l|ccc|ccc}
Method & $\delta_1 \uparrow$ & $\delta_2 \uparrow$ & $\delta_3 \uparrow$ & REL $\downarrow$ & RMS $\downarrow$ & $\log _{10} \downarrow$ \\
\hline
Chen et al. & 0.757 & 0.943 & 0.984 & 0.166 & 0.494 & 0.071 \\
Yin et al.~\cite{yin2019enforcing}  & 0.696 & 0.912 & 0.973 & 0.183 & 0.541 & 0.082 \\
BTS~\cite{lee2019big}  & 0.740 & 0.933 & 0.980 & 0.172 & 0.515 & 0.075 \\
AdaBins~\cite{bhat2021adabins}  & 0.771 & 0.944 & 0.983 & 0.159 & 0.476 & 0.068 \\
DepthFormer~\cite{li2022depthformer} & \underline{0.815} & \underline{0.970} & \underline{0.993} & \underline{0.137} & \underline{0.408} & \underline{0.059} \\
\rowcolor{gray!10} 
\ours (step 3) & \textbf{0.825} & \textbf{0.973} & \textbf{0.994} & \textbf{0.128} & \textbf{0.397} & \textbf{0.056} \\
\end{tabular}
\vspace{0.5em}
\caption{\textbf{Depth estimation on the SUN RGB-D dataset.}
We report the result of the model trained on the NYU-DepthV2 dataset and tested on the SUN RGB-D dataset without fine-tuning.}
\label{tab:exp:depth:sun}
\end{table}

\paragraph{NYU-DepthV2.}
The NYU-DepthV2 is an indoor scene dataset that consists of RGB and depth images captured at a resolution of 640$\times$480 pixels. 
The dataset contains over 1,449 pairs of aligned indoor scenes, captured from 464 different indoor areas.
We train \ours using image pairs with a resolution of 320$\times$240 and with varying depths up to approximately 10 meters.
Following previous work, we evaluate the results on the predefined center cropping by \cite{eigen2014depth}.
To be fair, all experimental configurations were aligned with the previous method \cite{li2022depthformer}.

\paragraph{SUN RGB-D.}
We use this dataset \cite{song2015sun} to evaluate generalization. 
To be specific, we assess the performance of our NYU pre-trained models on the official test set, which includes 5,050 images, without any additional fine-tuning. 
The maximum depth is restricted to 10 meters. 
Please note that this dataset is solely intended for evaluation purposes and is not utilized for training.

\begin{table}[t!]
\centering
\footnotesize
\setlength{\tabcolsep}{1.25mm}
\begin{tabular}{l|ccc|ccc}
Method & $\delta_1 \uparrow$ & $\delta_2 \uparrow$ & $\delta_3 \uparrow$ & REL $\downarrow$ & RMSE $\downarrow$ & $\log _{10} \downarrow$  \\
\hline 
StructDepth~\cite{li2021structdepth} & 0.817 & 0.955 & 0.988 & 0.140 & 0.534 & 0.060  \\
MonoIndoor~\cite{ji2021monoindoor} & 0.823 & 0.958 & 0.989 & 0.134 & 0.526 & -  \\
DORN~\cite{fu2018deep} & 0.828 & 0.965 & 0.992 & 0.115 & 0.509 & 0.051  \\
BTS~\cite{lee2019big} & 0.885 & 0.978 & 0.994 & 0.110 & 0.392 & 0.047  \\
DAV~\cite{huynh2020guiding} & 0.882 & 0.980 & 0.996 & 0.108 & 0.412 & -  \\
TransDepth~\cite{yang2021transformer} & 0.900 & 0.983 & 0.996 & 0.106 & 0.365 & 0.045  \\
DPT-Hybrid~\cite{ranftl2021vision} & \underline{0.904} & 0.988 & \textbf{0.998} & 0.110 & 0.357 & 0.045  \\
AdaBins~\cite{bhat2021adabins} & 0.903 & 0.984 & \underline{0.997} & 0.103 & 0.364 & 0.044  \\
DepthFormer~\cite{li2022depthformer} &  \textbf{0.921} & \underline{0.989} & \textbf{0.998} & \underline{0.096} & \underline{0.339} & \underline{0.041}  \\
\rowcolor{gray!10} 
\ours (step 3) & \textbf{0.921} & \textbf{0.990} & \textbf{0.998} &\textbf{0.094} & \textbf{0.329} & \textbf{0.040}   \\ 
\end{tabular}
\vspace{0.5em}
\caption{\textbf{Depth estimation on the NYU-DepthV2 val set.}
We report the performance of \ours with 3 diffusion steps.
The best and second-best results are bolded or underlined, respectively. ↓ means lower is better, and ↑ means higher is better.}
\label{tab:exp:depth:nyu}
\end{table}

\section{Experimental Results}
In \cref{tab:exp:depth:nyu}, we provide the depth estimation performance of DDP on the NYU-V2 dataset, in addition, in \cref{tab:exp:depth:sun}, we provide the generalization performance results of DDP on the SUN-RGBD dataset.

\label{supp:experiment}

\section{Visualization}
\label{supp:visualization}

\begin{figure*}[h]
    \centering
    \includegraphics[width=\linewidth]{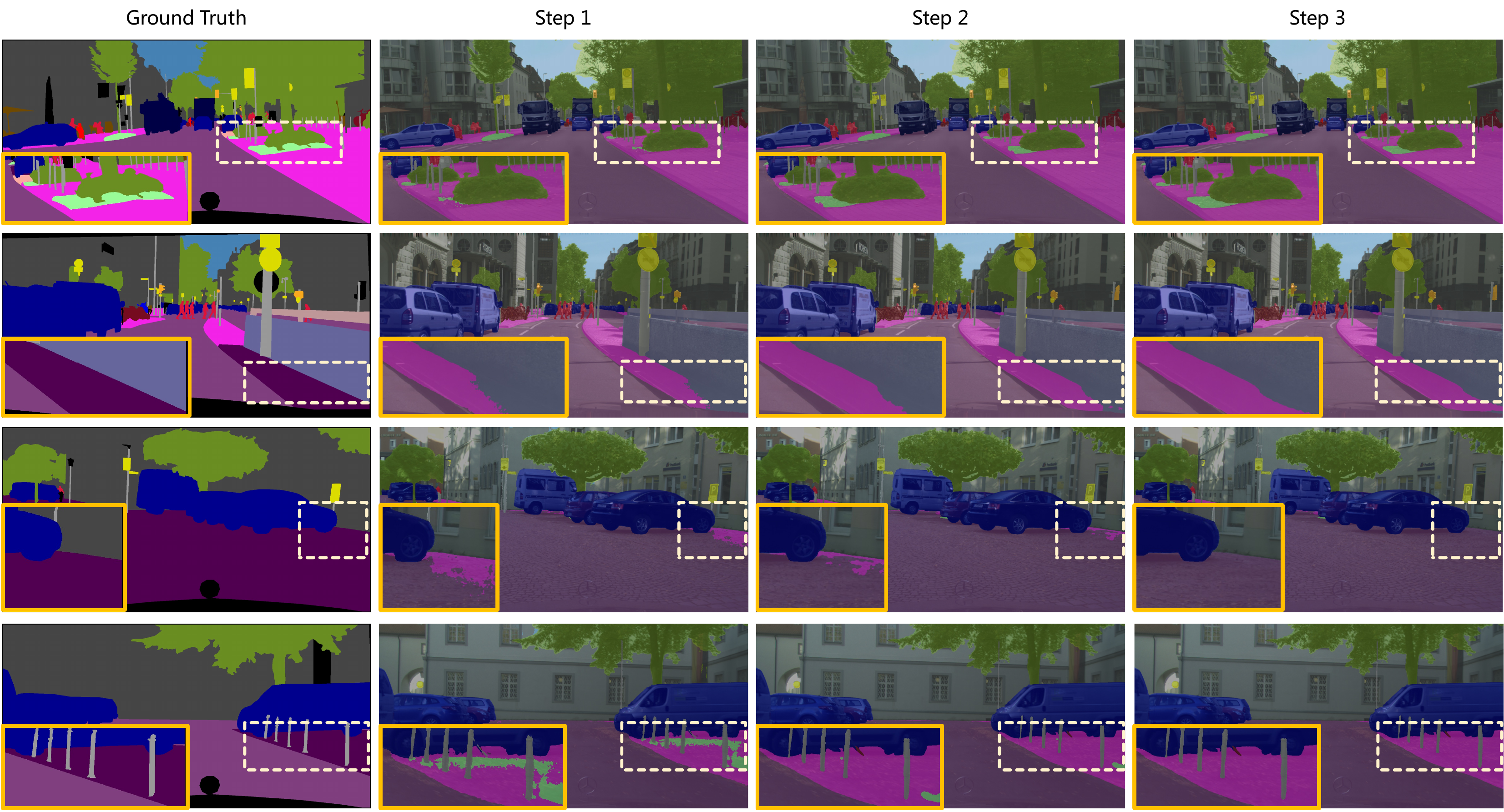}
    \caption{\textbf{Visualization of multiple inference on Cityscapes val set.}
     }
    \label{fig:vis:city_multi_steps}
\end{figure*}

\begin{figure*}[h]
    \centering
    \includegraphics[width=\linewidth]{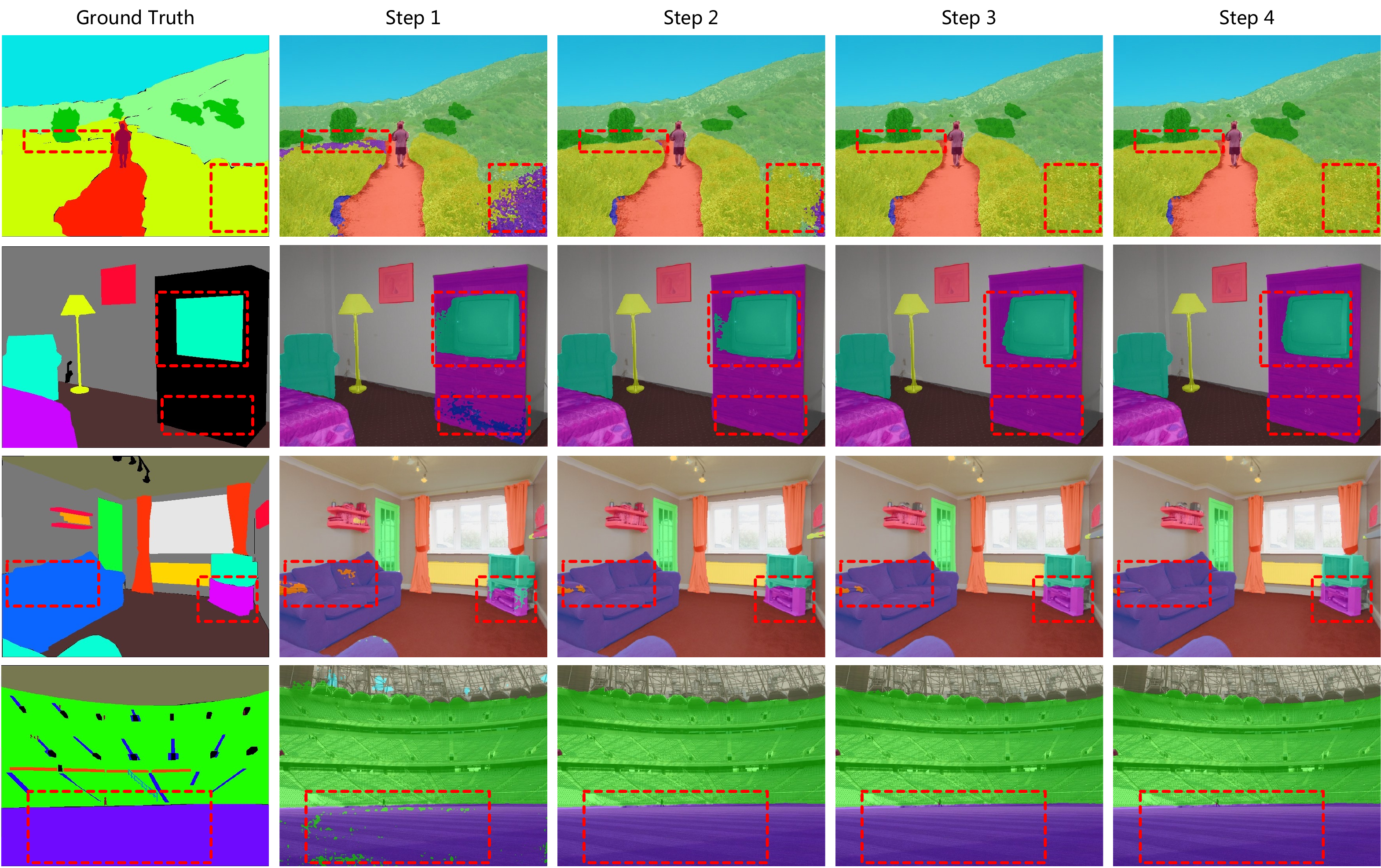}
    \caption{\textbf{Visualization of multiple inference on ADE20K val set.}
     }
    \label{fig:vis:ade_multi_steps}
\end{figure*}

\begin{figure*}[h]
    \centering
    \includegraphics[width=\linewidth]{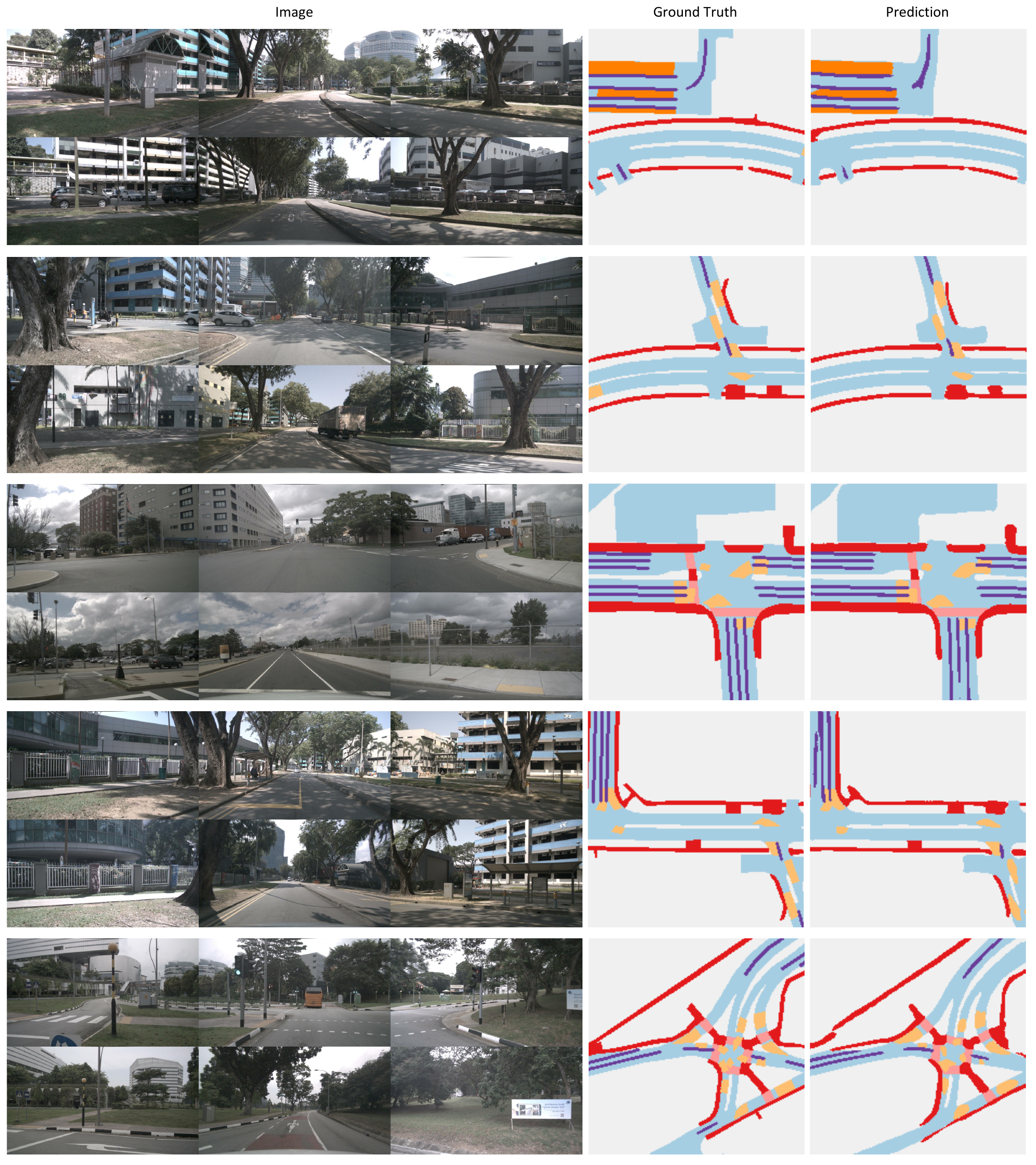}
    \caption{\textbf{Visualization of predicted BEV map segmentation results on nuScenes val set.}
     }
    \label{fig:vis:bev}
\end{figure*}

\begin{figure*}[h]
    \centering
    \includegraphics[width=\linewidth]{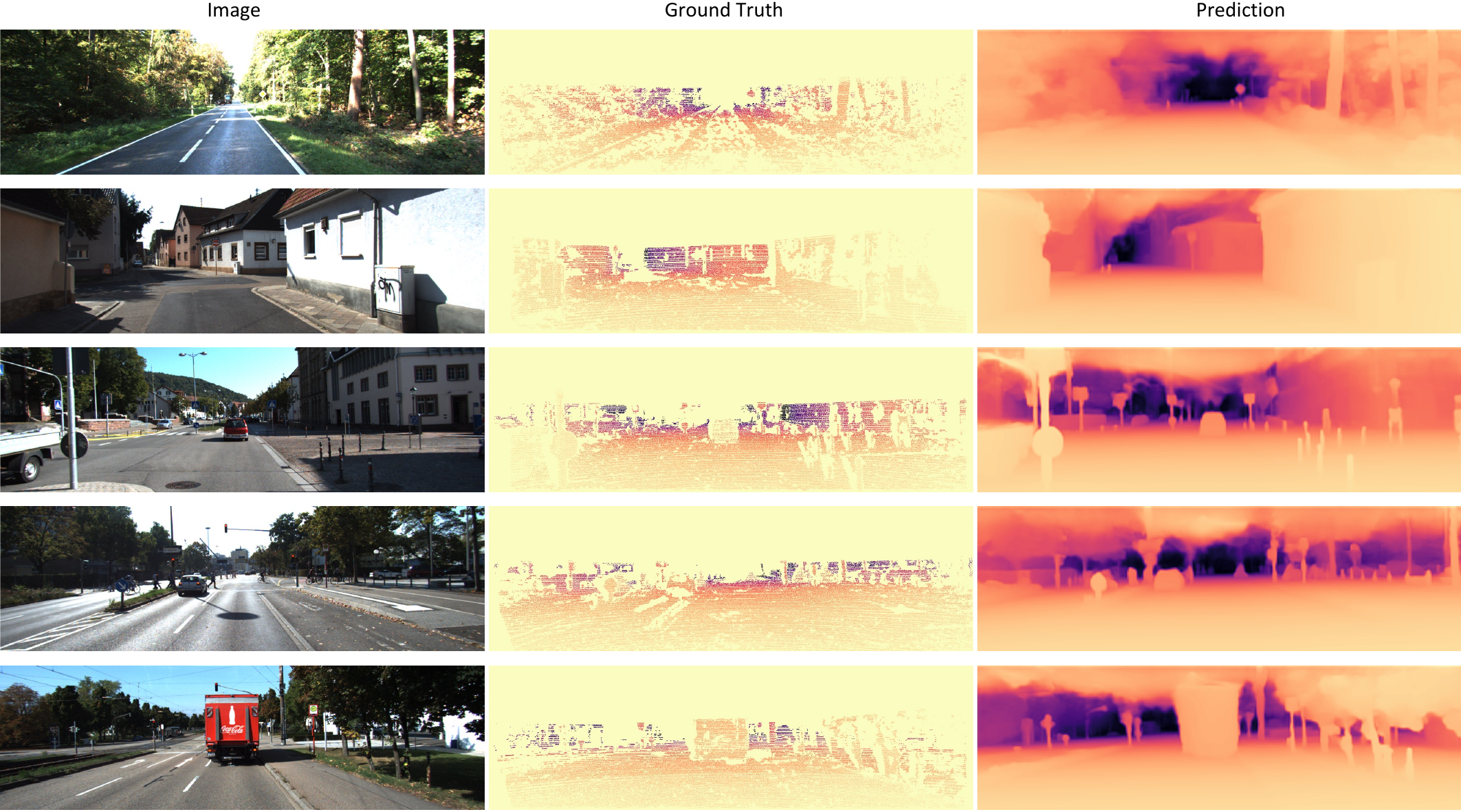}
    \caption{\textbf{Visualization of predicted depth estimation results on KITTI val set.}
     }
    \label{fig:vis:kitti}
\end{figure*}

\begin{figure*}[h]
    \centering
    \includegraphics[width=\linewidth]{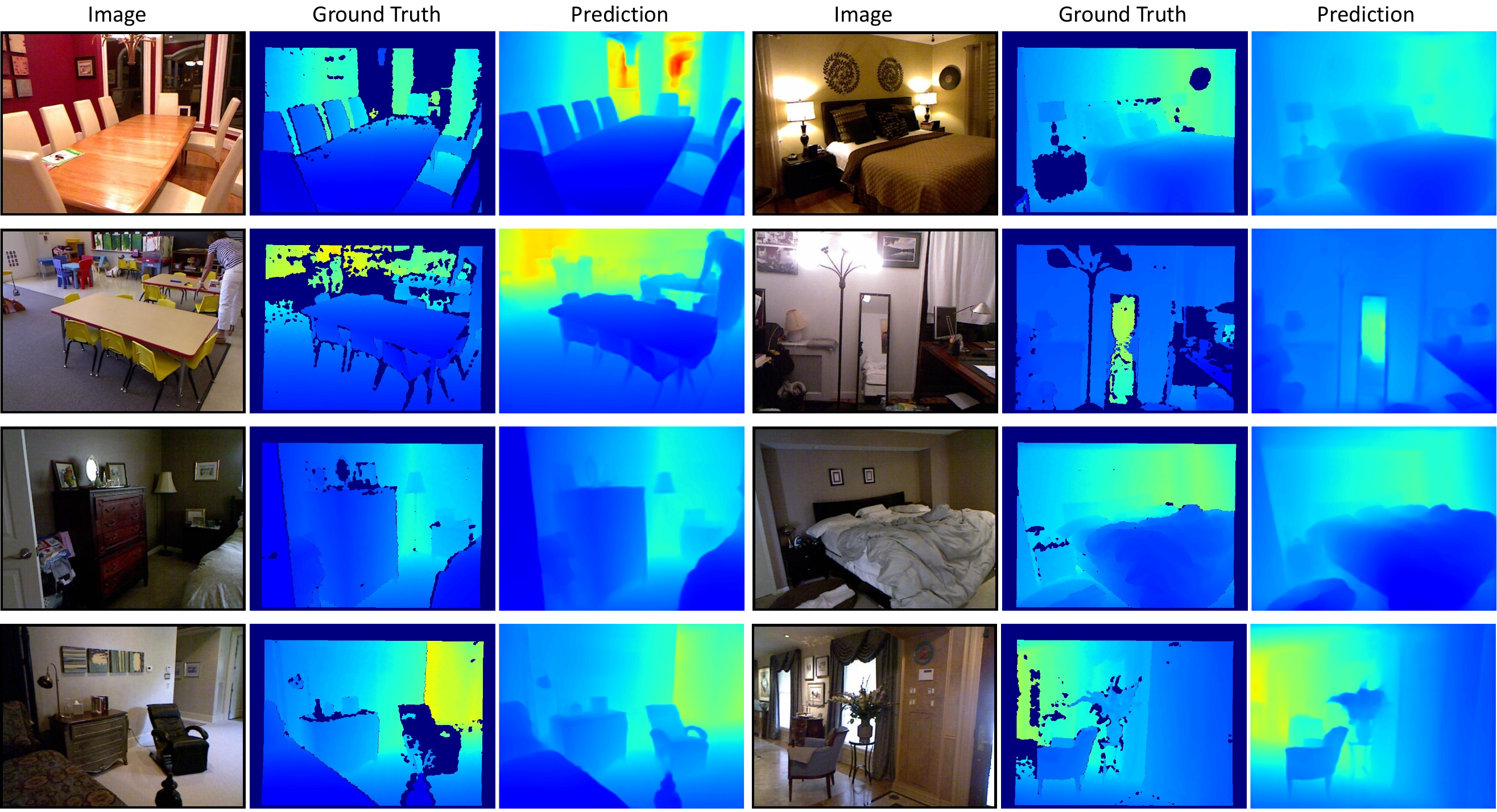}
    \caption{\textbf{Visualization of predicted depth estimation results on NYU-DepthV2 val set.}
     }
    \label{fig:vis:nyu}
\end{figure*}

\cref{fig:vis:city_multi_steps} and \cref{fig:vis:ade_multi_steps} visualize the ``multiple inference" property of \ours on the validation sets of Cityscapes and ADE20K, respectively.
These inference trajectories show that DDP can enhance its performance continuously and produce smoother segmentation maps by using more sampling steps.
\cref{fig:vis:bev} presents the BEV map segmentation results of \ours (step 3) with the ground truths and multi-view images.
\cref{fig:vis:kitti} and \cref{fig:vis:nyu} compare the generated depth estimation results of \ours (step 3) with the ground truths on the validation sets of KITTI and NYU-DepthV2, respectively. 
These results indicate that our method can be easily generalized to most dense prediction tasks.

\newpage

\newpage

\newpage

\newpage

\newpage

\newpage

\section{More Applications}
\subsection{Combine DDP with ControlNet}
\paragraph{Setup.}
It has been found that compared to the previous single-shot model, DDP can achieve more continuous and semantic consistency prediction results. To demonstrate the benefits of this pixel clustering property, we combined DDP with the recently popular segmentation mask condition generation model: ControlNet. We followed the official implementation of ControlNet for all hyperparameters, including input resolution and DDIM sampling steps.

\paragraph{Implementation}
ControlNet~\cite{zhang2023adding} improves upon the original Stable Diffusion (SD) model by adding extra conditions, which is done by incorporating a conditioning network.
In the mask-conditional ControlNet, the map generated by the segmentation model is used as input for image synthesis.
The original segmentation model was adopted from Uniformaer-S~\cite{li2022uniformer} with UperNetHead, which has 52M parameters and achieves 47.6 mIoU (ss) on the ADE20K dataset.
To make a fair comparison, we replaced the original segmentation model in the mask-conditional ControlNet with DDP using the Swin-T backbone, which has 40M parameters and achieves 47.0 mIoU (ss) on the ADE20K dataset.
Note that all results were obtained with the default prompt.

\begin{figure*}[h]
    \centering
    \includegraphics[width=0.775\linewidth]{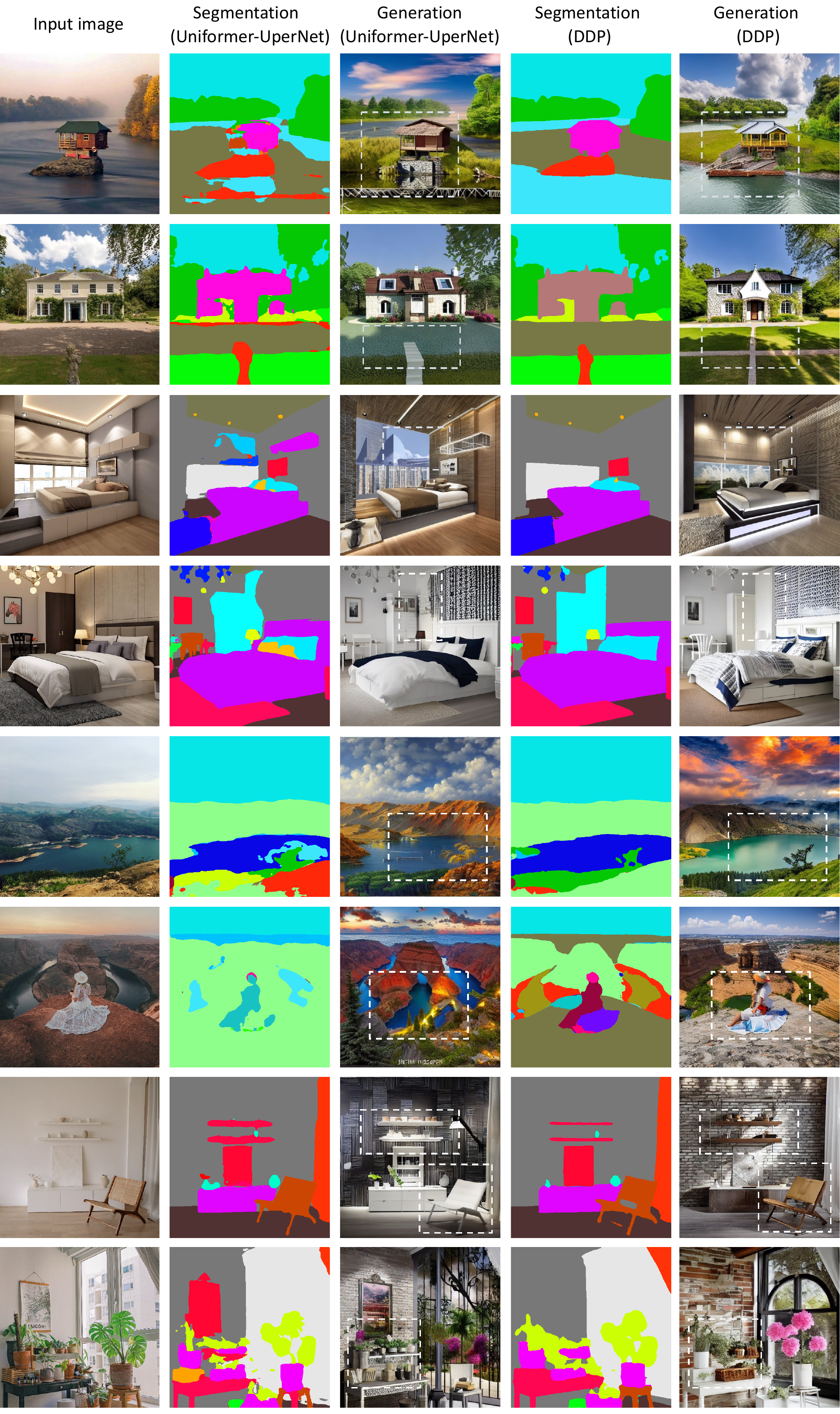}
    \caption{\textbf{Control Stable Diffusion with Semantic Map}, the Uniformer-UnperNet, and DDP segmentation models are used to predict segmentation maps as condition input. All results were achieved using the default prompt.
     }
    \label{fig:vis:ddp_control}
\end{figure*}

\paragraph{Results}
We select images from the PEXEL website \url{https://www.pexels.com/} for testing in different scenarios.
The results from the original ControlNet and the combination of DDP with ControlNet are shown in \cref{fig:vis:ddp_control}.
ControlNet is designed to achieve fine-grained, controllable image generation, our experiments show that DDP can produce more consistent results and has advantages in various scenarios. Moreover, when combined with DDP, ControlNet produces visually satisfying and well-composed results, surpassing those of the original ControlNet.
Our experimental results suggest that DDP has great potential to improve cooperation with other types of foundation models.

\newpage
\clearpage

\end{document}